\documentclass[10pt,twocolumn,letterpaper]{article}

 \usepackage{cvpr}              

\newcommand{\rawdet}{{\sc RAWDet-7}}

\newcommand{\method}{{\sc $\gamma$-scaling}}

\title{\rawdet: A Multi-Scenario Benchmark for Object Detection \\ and Description on Quantized RAW Images}
\definecolor{cvprblue}{rgb}{0.21,0.49,0.74}
\usepackage[pagebackref,breaklinks,colorlinks,allcolors=cvprblue]{hyperref}

\usepackage[toc,page]{appendix}
\usepackage{minitoc}

\usepackage{titletoc}

\usepackage[dvipsnames]{xcolor}

\usepackage{booktabs}
\usepackage{algpseudocode}
\usepackage{algorithm}
\usepackage{multirow}
\usepackage{multicol}
\usepackage{mathtools}
\usepackage{xcolor}
\usepackage{ragged2e}
\usepackage[export]{adjustbox}
\usepackage{soul}
\usepackage{enumitem}
\usepackage[usestackEOL]{stackengine}
\usepackage[most]{tcolorbox}
\tcbuselibrary{listings}

\newtcblisting{promptbox}[1][]{
  listing only,
  colback=blue!1,
  colframe=blue!60!black,
  title=#1,
  boxrule=0.8pt,
  arc=2pt,
  left=6pt,
  right=6pt,
  top=6pt,
  bottom=6pt,
  listing options={
    basicstyle=\ttfamily\small,
    breaklines=true,
    columns=fullflexible,
    keepspaces=true
  }
}

\usepackage[capitalize]{cleveref}
\crefname{section}{Sec.}{Secs.}
\Crefname{section}{Section}{Sections}
\Crefname{table}{Table}{Tables}
\crefname{table}{Tab.}{Tabs.}
\definecolor{cadmiumgreen}{rgb}{0.0, 0.42, 0.24}
\definecolor{custom}{cmyk}{0.1,0.48,0.49,0.2}
\definecolor{OliveGreen}{cmyk}{0.64,0,0.95,0.40}
\definecolor{new}{rgb}{0.81,0.05,0.9}
\definecolor{BrickRed}{rgb}{0.81,0.1,0.1}
\definecolor{RoyalBlue}{rgb}{0.2,0.2,0.75}

   \usepackage{xspace}
   \usepackage[normalem]{ulem}
   \usepackage{bm}

\makeatletter
\DeclareRobustCommand\onedot{\futurelet\@let@token\@onedot}
\def\@onedot{\ifx\@let@token.\else.\null\fi\xspace}

\def\eg{e.g\onedot} 
\def\ie{i.e\onedot}

\makeatother

\def\clap#1{\hbox to 0pt{\hss #1\hss}}%
\def\initials#1{\protect\clap{\protect\smash{\protect\raisebox{1.4ex}{\protect\tiny{\protect\textsf{\protect\textit{#1}}}}}}}%
\makeatletter
\newcommand{\EDIT}[4][]{\protect\@ifundefined{hidecomments}{%
  \protect\strut{\color{#3}{\hspace{0pt}\initials{#2}\protect\sout{#1}{~#4}}}%
  }{#4}}
\newcommand{\NOTEboxed}[3]{\protect\@ifundefined{hidecomments}{%
  {\begin{center}\fbox{\parbox{0.97\linewidth}{\protect\EDIT{#1}{#2}{#3}}}\end{center}}
  }{}}
\newcommand{\COMM}[3]{\protect\@ifundefined{hidecomments}{%
  {\protect\EDIT{#1}{#2}{#3}}%
  }{}}
\newcommand{\DefAuthor}[2] 
{%
  \expandafter\newcommand\csname #1edit\endcsname[2][]{\protect\EDIT[##1]{#1}{#2}{##2}}
  \expandafter\newcommand\csname #1\endcsname[1]{\protect\COMM{#1}{#2}{[##1]}}
  \expandafter\newcommand\csname #1boxed\endcsname[1]{\protect\NOTEboxed{#1}{#2}{##1}}
}
\definecolor{dfltgreen}       {rgb}{0.0,0.5,0.0}
\definecolor{dfltred}         {rgb}{0.7,0.0,0.0}
\newcommand{\REVadd}[1]{\protect\@ifundefined{hidecomments}{%
  \strut{\color{dfltgreen}{#1}}}{#1}}
\newcommand{\REVedit}[2][]{\protect\@ifundefined{hidecomments}{%
  \strut{\color{dfltred}{\protect\sout{#1}}\color{dfltgreen}{~#2}}}%
  {#2}}
\makeatother

\definecolor{dkgreen}       {rgb}{0.0,0.5,0.0}
\definecolor{dkblue}        {rgb}{0.0,0.0,0.7}
\definecolor{dkcyan}        {rgb}{0.0,0.5,0.5}
\definecolor{dkmagenta}     {rgb}{0.5,0.0,0.5}
\DefAuthor{MK}{dkmagenta} 
\DefAuthor{MM}{dkgreen} 
\DefAuthor{VG}{dkblue} 
\DefAuthor{MF}{dkcyan} 
\DefAuthor{SA}{orange} 
\DefAuthor{todo}{red}

\usepackage{pifont}
\def\confName{CVPR}
\def\confYear{2026}

\author{
Mishal Fatima\textsuperscript{1}\thanks{These authors contributed equally to this work. \\Emails: mishal.fatima@uni-mannheim.de, shashank.agnihotri@uni-mannheim.de},
Shashank Agnihotri\textsuperscript{1}$^*$,
Kanchana Vaishnavi Gandikota\textsuperscript{2},\\
Michael Moeller\textsuperscript{2},
Margret Keuper\textsuperscript{1,3} \\[1ex]
\textsuperscript{1}University of Mannheim, Germany 
\textsuperscript{2}University of Siegen, Germany \\
\textsuperscript{3}Max Planck Institute for Informatics, Saarland Informatics Campus, Germany
}

\begin{document}
\maketitle
\begin{abstract}
Most vision models are trained on RGB images processed through ISP pipelines optimized for human perception, which can discard sensor-level information useful for machine reasoning. RAW images preserve unprocessed scene data, enabling models to leverage richer cues for both object detection and object description, capturing fine-grained details, spatial relationships, and contextual information often lost in processed images. To support research in this domain, we introduce \rawdet{}, a large-scale dataset of $\sim$25k training and 7.6k test RAW images collected across diverse cameras, lighting conditions, and environments, densely annotated for seven object categories following MS-COCO and LVIS conventions. In addition, we provide object-level descriptions derived from the corresponding high-resolution sRGB images, facilitating the study of object-level information preservation under RAW image processing and low-bit quantization. The dataset allows evaluation under simulated 4-bit, 6-bit, and 8-bit quantization, reflecting realistic sensor constraints, and provides a benchmark for studying detection performance, description quality \& detail, and generalization in low-bit RAW image processing. Dataset \& code upon acceptance.
\end{abstract}

\section{Introduction}
\label{sec:intro}
\begin{figure*}
    \centering
    \includegraphics[width=0.9\linewidth]{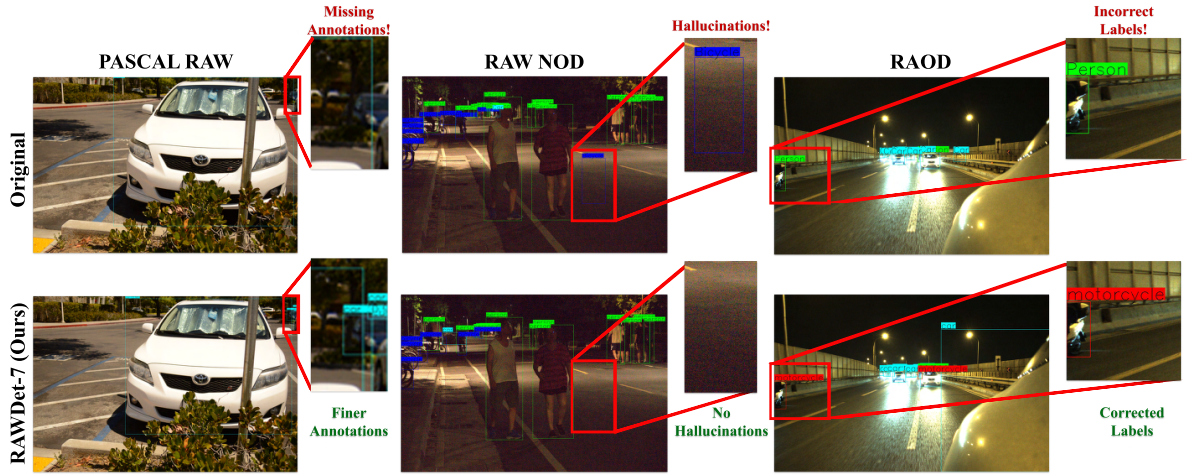}
    \caption{Comparing ground truth annotations provided in the original datasets and the new ones proposed in \rawdet{}. Our proposed annotations are more fine-grained, as seen for PASCAL RAW, which originally annotated only one instance of the cars in the image; we now annotate all the other instances of cars in the image (with a 20\% overlap threshold \ie at least 20\% area of the bounding box should be non-overlapping). RAW NOD (-Nikon and -Sony), RAOD (-Day and -Night) original annotations contain hallucinations as seen here for RAW NOD, which hallucinates a bicycle in the center right of the frame. Original annotations even contain some misclassifications, as seen for RAOD, which misclassified a motorcycle as a person. \rawdet{} \textbf{(bottom) overcomes these drawbacks.}}
    \label{fig:teaser}
    \vspace{-1em}
\end{figure*}

Most vision pipelines operate on standard RGB (sRGB) images with 8-bit depth. During image capture, however, cameras record visual information in a much higher bit-depth ($\sim$24-bit) RAW format, which is then converted to sRGB by an image signal processor (ISP). This conversion involves a sequence of steps, including black-level subtraction, demosaicking, denoising, white balancing, gamma correction, color correction, and compression, and is usually conceived to produce a visually pleasing 8-bit color image.
However, this photography-oriented imaging pipeline is not optimized for downstream vision tasks. 
Compared with processed sRGB images, RAW images contain more precise sensor measurements, preserving the full bit depth and dynamic range. Leveraging RAW data for downstream tasks can therefore improve performance \cite{omid2014pascalraw,xu2023RAODdataset}. RAW data, however, varies significantly across camera sensors, each of which may use different bit depths and ISP pipelines to produce sRGB images. As a result, for specialized applications~\cite{klinghoffer2022physics,sommerhoff2023differentiable,agarwal2025directimageclassificationfourier}, jointly optimizing the imaging hardware or ISP pipeline together with the downstream task can be beneficial.
Despite these potential advantages, progress in using RAW data and end-to-end learned, task-specific image processing remains constrained by the limited availability of large-scale, well-annotated RAW datasets.


To address this, we introduce \rawdet, a comprehensive dataset for object detection and description on RAW images. 
We chose the task of object detection as it requires both identifying and localizing features associated with objects of specific classes within images. 
Furthermore, we provide a curated subset of sRGB images accompanied by object descriptions, specifically designed to assist researchers in evaluating the level of detail preserved in processed RAW images.
The proposed \rawdet{} has a larger sample size than all previous datasets, comprising $\sim$25k training and $\sim$7.6k test images obtained by consolidating four different datasets \cite{omid2014pascalraw,xu2023RAODdataset,morawski2022genisp,ignatov2020replacing} across varied lighting conditions, camera types, and environments. 
As many prior datasets have only a few object classes, typically covering only large prominent objects, we derive more fine-grained labels by relabeling the consolidated dataset using a large foundation model with subsequent human validation on the sRGB versions of the RAW images to annotate 7 object categories, that align with community accepted naming conventions from MS-COCO~\cite{ms-coco} and LVIS~\cite{gupta2019lvis} datasets. 
As a result, the proposed dataset provides denser annotations, both in terms of increasing the number of object categories and improving the annotation of small and occluded instances, which significantly enhances the utility of the proposed dataset, \rawdet{}, in developing robust detection models that generalize to real-world scenarios.
This can be observed in \cref{fig:teaser}, \rawdet{} has significantly richer and corrected annotations than previous datasets for the same images. 
While \rawdet{} comprises full-precision RAW images, reducing the bit depth has shown benefits in memory consumption and significant power gains~\cite{power_gain_low_bit_2021}. 
However, there is limited knowledge on the performance of complex downstream vision methods on heavily quantized RAW input.
Thus, we benchmark the performance of object detection models under simulated low-bit quantization, in addition to sRGB images.

 We consider low-bit quantized inputs obtained using different information processing methods: linear scaling,  logarithmic scaling~\cite{log_quant_Buckler_2017_ICCV}, learnable $\gamma$ scaling~\cite{ljungbergh2023raw}, and a combination of logarithmic and learnable $\gamma$ scaling~\cite{fatima2025gamma}. 
 Through analytical benchmarking of object detection methods low-bit quantized RAW inputs, we demonstrate that models trained on the proposed \rawdet{} have an improved object detection performance over models trained on the individual subsets across different quantization levels, with more robust detections on small and occluded instances, demonstrating the benefit of the proposed dataset. 
 Moreover, note that all prominent object detection methods proposed, including every design decision made, were optimized for sRGB input.
 We show in our benchmarking that, despite this obvious bias, object detection methods trained on RAW input, especially low-bit RAW input, with input scaling can perform at-par with (sometimes even surpass) their counterparts trained on the same amount of sRGB input. We also evaluate the fidelity of processed RAW images by comparing the object descriptions obtained from different scaling variants with the ground truth captions generated from high-resolution sRGB images. Our results suggest that descriptions generated from processed RAW images exhibit greater similarity to those from high-resolution sRGB images and provide more accurate explanations of objects compared than captions from linear RAW images.

The contributions of this work are as follows:
\begin{itemize}
\item We introduce a comprehensive dataset for object detection on RAW images, designed to support research in low-bit quantized sensing from RAW data.
\item We analytically benchmark object detection models on RAW images quantized to different bit depths using four input scaling methods.
\item We demonstrate that RAW images quantized to as low
as 4 bits can perform on-par or better than the models trained on standard 8-bit RGB images.
\item We demonstrate the utility of our large-scale dataset in improving the performance of object detection models across different sensors, lighting conditions, and bit depths compared to the individual datasets.
\item By providing object-level descriptions for a subset of images derived from high-resolution sRGB images, we create a ground truth reference that can be used to evaluate captions generated from other image representations.
\item Lastly, we show that the observed benefits of \rawdet{} and input scaling methods are not limited to traditional object detection methods but also extend to recent large foundational model-based object detection methods. 
\end{itemize}

\section{Related Work}
\label{sec:related}
In the following, we discuss prior works towards object detection on RAW images, and relevant RAW image datasets that align closely with the proposed \rawdet{}.

\vspace{0.1cm}
\noindent\textbf{Object detection. }
We rely on established object detection architectures to benchmark and analyze performance on RAW and quantized RAW image formats. \textbf{Faster R-CNN}~\cite{faster_rcnn} serves as a strong two-stage baseline, leveraging region proposal networks for object localization. 
To assess performance under a one-stage paradigm, we include \textbf{RetinaNet}~\cite{lin2017focal}, which uses focal loss to address class imbalance, and \textbf{PAA}~\cite{kim2020probabilistic_paa}, which adaptively selects positive samples through probabilistic anchor assignment. 
These models are widely used in both academic and industrial settings due to their strong performance and robustness across a variety of detection benchmarks. 
Finally, to evaluate whether our observations generalize to more recent and large-scale detection architectures, we also include \textbf{MM-Grounding DINO}~\cite{mm_grounding_dino}, a foundation model for open-set object detection that we employ in a frozen setting. 
We use this diverse set of methods to enable a broad and representative analysis of detection performance in the RAW image domain. 

\vspace{0.1cm}
\noindent\textbf{RAW Imaging Datasets. }Most existing datasets and models for object detection are developed on RGB images produced by ISP pipelines optimized for human perception. In contrast, RAW images preserve richer scene information and higher dynamic range, which may benefit high-level tasks such as detection~\cite{ljungbergh2023raw,xu2023RAODdataset,wu2024dense}. However, large-scale annotated RAW datasets remain scarce. Existing detection datasets like RAOD~\cite{xu2023RAODdataset}, NOD~\cite{morawski2022genisp}, MultiRAW~\cite{li2024efficient_multiraw} and PASCAL RAW~\cite{omid2014pascalraw} are limited in either the number of images, object classes, or annotation quality, PASCAL RAW annotates only large objects, NOD and RAOD contain hallucinated or noisy labels, and all three cover a narrow range of conditions. The Zurich dataset ~\cite{ignatov2020replacing}, originally designed for RAW-to-RGB ISP learning, lacks detection labels and is re-annotated in our work. Other RAW datasets~\cite{ignatov2020replacing,8478390,dang2015raise} focus on ISP or forensics, not object detection.

To address these limitations, prior works propose synthetic RAW data generation~\cite{10415533,ISP-Teacher} or task-adaptive ISP modules~\cite{cui2024raw,diamond2021dirty,mosleh2020hardware,robidoux2021end} for improving detection. Some methods~\cite{log_quant_Buckler_2017_ICCV,xu2023RAODdataset} evaluate the effect of low-bit quantization on recognition but do not optimize quantization-aware pipelines, while others~\cite{ljungbergh2023raw,liu2024learnable} jointly learn simplified ISPs with detection models. In contrast, our proposed dataset, \rawdet{}, unifies and re-annotates existing datasets with fine-grained labels for seven categories across varied lighting, sensor types, and scenes, and enables controlled benchmarking under realistic 4-bit, 6-bit, and 8-bit quantization. For evaluation, we adopt standard detectors including Faster R-CNN~\cite{faster_rcnn}, RetinaNet~\cite{lin2017focal}, and PAA~\cite{kim2020probabilistic_paa}, as well as MM-Grounding DINO~\cite{mm_grounding_dino}, demonstrating that our findings generalize across architectures and detection paradigms.

\vspace{0.1cm}
\noindent\textbf{Object Description. } Recent works have explored set-of-mark and visual prompting strategies to improve region-level grounding and object-centric captioning in large vision-language models \cite{yang2023set_of_marks,wan2024contrastive_set_of_marks_2,lei-etal-2025-scaffolding_set_of_marks_3,cai2024vipllava_set_of_marks_4,yang2023fine_set_of_marks_5}. 
Set-of-Mark prompting overlays alphanumeric marks and segmentation-derived regions so that a VLM can reason about specific objects and their relations directly on the image~\cite{yang2023set_of_marks}. 
\cite{wan2024contrastive_set_of_marks_2} instead compares predictions on images with highlighted versus blacked-out regions to reduce prior biases and better follow region prompts. 
\cite{lei-etal-2025-scaffolding_set_of_marks_3} introduces a coordinate grid on the image together with textual references to enhance spatial reasoning and vision-language coordination. 
ViP-LLaVA~\cite{cai2024vipllava_set_of_marks_4} trains multimodal models to interpret arbitrary user-drawn prompts such as boxes, arrows, and scribbles as region selectors, while \cite{yang2023fine_set_of_marks_5} uses precise segmentation masks and blur-outside-mask operations to focus the model on a target instance. 
In \rawdet{}, we adopt a set-of-marks interface and additionally provide a high-level object class for each numeric mark when prompting the VLM, allowing us to study how well detailed object descriptions are preserved across different quantizations.


\section{Proposed Dataset \rawdet}
 \rawdet{} provides high quality annotations for object detection as well as for object description. The following sections provide the details of the annotation process. 
\begin{table*}[htbp]
\centering
\caption{Comparison of \rawdet{} with other Datasets. Please refer to the Appendix for a more detailed comparison.}
\label{tab:dataset_comparison}
\footnotesize
\begin{tabular}{l@{\hspace{3em}}c@{\hspace{3em}}c@{\hspace{3em}}c@{\hspace{3em}}c@{\hspace{3em}}c@{\hspace{3em}}c}
\toprule
Dataset & Train Images & Test Images & \# Classes & Bit Depth & Scenarios & \# Sensors\\
\midrule
PASCAL RAW & 2128 & 2130 & 3 & 12-bit & Day & 1\\
RAW NOD (ROD) & 6692 & 1671 & 3 & 14-bit & Night & 2 \\
RAOD & 16089 & 4000 & 5 & 24-bit & Day and Night & 1\\
MultiRAW & 5154 &  2315 & 10 & 10, 12, 14, 24-bit & Day and Night & 5\\
\rawdet{} \textbf{(Ours)} & 24864 & 7688 & 7 & 10, 12, 14, 24-bit & Day and Night & 5 \\
\bottomrule
\end{tabular}
\end{table*}
\vspace{-1em}

\subsection{Detection Annotations}
As discussed in \cref{sec:intro},  RAW images have been underutilized in research due to a lack of large-scale, high-quality benchmarks and the practical challenges of managing RAW data. 
Although several RAW object detection datasets \cite{omid2014pascalraw,morawski2022genisp,xu2023RAODdataset} have been proposed in the literature, we discuss in \cref{sec:related} that these datasets have practical limitations. 
For instance, PASCAL RAW \cite{omid2014pascalraw} and NOD \cite{morawski2022genisp} focus on only a few coarse object classes, limiting object categories to `Car', `Person', and `Bicycle'.  
In \cref{fig:teaser} we observed several missing annotations even among these classes for small or partially occluded people and vehicles. 
NOD dataset~\cite{morawski2022genisp} suffers from hallucinations and missing annotations, \eg, one could observe in \cref{fig:teaser} that a motorcycle (`vehicle' in the original category) from the RAOD~\cite{xu2023RAODdataset} dataset is being mislabeled as a `Person'.
Apart from the errors and missing annotations,  sensor-specific differences in bit depth and lighting conditions introduce significant variability, captured only partially by previous datasets.
Notably, MultiRAW~\cite{li2024efficient_multiraw} attempted increasing the number of classes and scenarios, but it has contradictory classes like `rider' and `person' and too few total samples making the problem complex.

To address these gaps, we introduce \rawdet, a comprehensive benchmark dataset for object detection on RAW images. 
Comprising over 32k RAW images of which there are ~25k images for training and 7.6k for testing, \rawdet{} consolidates and standardizes four existing datasets: PASCAL RAW~\cite{omid2014pascalraw} (12-bit depth), RAOD~\cite{xu2023RAODdataset} (24-bit depth), covering day and night scenes in HDR format, Zurich RAW dataset \cite{ignatov2020replacing} (10-bit depth), NOD dataset~\cite{morawski2022genisp} (14-bit depth) containing data from Sony and Nikon sensors. 
These sources cover a wide range of imaging conditions, including daytime, nighttime, and high dynamic range (HDR) scenes, as well as diverse hardware and sensor bit depths. 
We unify these disparate datasets under a consistent annotation scheme, thereby enabling controlled studies of model generalization across lighting, scene complexity, and sensor variability.  
A key contribution of \rawdet{} is its improved and standardized labeling. 
To remedy the deficiencies in annotations in the individual datasets, and to enable a uniform standardized annotation, we re-annotated all datasets using Grounded-DINO 1.5 \cite{liu2024grounding} (paid API version at: \url{https://deepdataspace.com/request_api}) for seven object categories: \emph{`Car'}, \emph{`Truck'}, \emph{`Tram'}, \emph{`Person'}, \emph{`Bicycle'}, \emph{`Motorcycle'}, and \emph{`Bus'} following MS-COCO and LVIS-style naming conventions, and further refined the annotations produced by Grounded-DINO 1.5 by selecting confident annotations above the confidence threshold of 0.8. 
The cut-off threshold was chosen after manual inspection of the annotation quality and accuracy over a large subset of \rawdet{}.
As a result, \rawdet{} has dense annotations with uniform, high-quality labeling standards across the combined dataset with improved labeling of small and occluded object instances. The statistics of individual object categories in the consolidated dataset, as well as the individual subsets, are provided in \cref{fig:dataset_statistics}. \cref{tab:sensor_statistic} lists the properties of individual subsets of \rawdet{}.

\subsection{Object Description}
While being precise in terms of localization, object detections only provide a coarse, class-level understanding of objects in a scene. 
To evaluate the level of detail preserved in processed RAW images, in particular on the highly relevant, annotated object categories, we propose to consider the task of \emph{object description}. 
To this end, we generate a set of high-quality, detailed object descriptions using the Gemini-2.5-Pro model \citep{comanici2025_gemini-2.5} for objects localized by the ground truth bounding box annotations from \rawdet{}. Specifically, we overlay, for a subset of 500 high-resolution sRGB images, visual marks on annotated objects, in line with the procedure outlined in \cite{yang2023set_of_marks}, using black, squared marks with colored numbers indicating the object ID. We then prompt Gemini-2.5-Pro to provide detailed descriptions for every object. To provide a reference, we prompt the model twice per image, thereby generating an upper bound on description agreement. 
Our full prompts and examples of object descriptions are provided in the appendix, along with the user study for dataset validation. 

For evaluation, we calculate the RegeX overlap and BLeU score between captions generated for the high-resolution sRGB image, and downsampled sRGB variants and differently processed RAW variants of different bit-depths, for the same set-of-marks.
Additionally, using Gemini-2.5-Flash-Lite as a judge (score 1-10), we calculate the semantic similarity (disregarding details), level of detail in caption (disregarding similarity), and the match in precise details of the pair of captions (sRGB and variant).

\section{Scaling Methods Used For Benchmarking}
\label{sec:methods:benchmarking}
One key challenge with using RAW images in machine learning is that their high precision and large size, make them computationally expensive to store and process. Quantization offers a practical solution by reducing bit depth, but it introduces new design choices around how to scale inputs before quantization. As an example use-case, in this section, we explore combinations of quantization with known input scaling methods, such as logarithmic and gamma mappings, to enable efficient and effective object detection directly on low-bit RAW inputs. Recording at lower bit-depth and skipping the ISP before feeding images to a downstream task model can provide power and memory gains~\cite{power_gain_low_bit_2021},
making such methods practical in scenarios like low-bandwidth or low-power consumption sensing.

We benchmark the performance of \rawdet{} using low-bit depth images obtained using the following scaling methods, where $\mathcal{X}$ with bit-depth $N$ is the input to the quantizer $\mathcal{Q}(\cdot)$, which provides output with  target bit depth $\hat{N}$:
 \begin{enumerate}[leftmargin=*]
 \item\textbf{Linear quantization} converts input into a quantized digital value using uniform linear steps as
 \begin{equation}
    \label{eqn:linear_quantization}
    \mathcal{Q}(\mathcal{X})= \left\lfloor{\mathcal{X}\textsubscript{norm}\cdot (2^\mathrm{\hat{N}}-1)}\right\rfloor,
\end{equation}
where, $\mathcal{X}\textsubscript{norm}=\left(\frac{\mathcal{X}}{2^\mathrm{N} - 1}\right)$ is the normalized version of $\mathcal{X}$.
This na\"ive approach of reducing the bit-depth often fails, motivating better methods for scaling.

\item\textbf{Logarithmic quantization} proposed by \cite{log_quant_Buckler_2017_ICCV,adaptive_log_pixel_sensor} non-linearly scales the dynamic range of $\mathcal{X}$ before quantization,
\begin{equation}
\mathcal{Q}(\mathcal{X}) = \left\lfloor{ \frac{\mathcal{X}_{\mathrm{log}} - \mathrm{min}(\mathcal{X}_{\mathrm{log}})}{\mathrm{max}(\mathcal{X}_{\mathrm{log}}) - \mathrm{min}(\mathcal{X}_{\mathrm{log}})} \cdot (2^\mathrm{\hat{N}}-1)}\right\rfloor ,
\label{eq:log_quant}
\end{equation}
where, $\mathcal{X}_{\mathrm{log}} = \log(\mathcal{X}+\epsilon)$, with $\epsilon$ is accounting for the mapping to issue values $\mathcal{X}_\mathrm{log}\geq 0$, i.e., \cite{log_quant_Buckler_2017_ICCV} use $\epsilon=1$.
\item\textbf{Quantization with Learnable  $\gamma$ -Scaling:}  $\gamma$ mapping, similar to logarithmic quantization, also compresses the dynamic range of the input, and we consider a learnable $\gamma$ mapping before quantization as 
\begin{equation}
\label{eqn:our_gamma_teaser}
\mathcal{Q}(\mathcal{X}, \gamma) =\left\lfloor \mathcal{X}_{\text{norm}}^{\gamma} \cdot \left(2^{\mathrm{\hat{N}}}-1\right)\right\rfloor,
\end{equation}
where $\mathcal{X}\textsubscript{norm}$ is the normalized version of $\mathcal{X}$ as defined earlier in linear quantization, and the parameter $\gamma$ is learned along with the parameters $\theta$ of the neural network. In addition to learning a single $\gamma$ for the dataset, we also evaluate conditioning $\gamma$ on the domain, such as lighting conditions or sensors.
$\mathcal{Q}(\mathcal{X}, \gamma)$ is hereby referred to as \method{}.

\item Since various non-linear mappings of input intensities are an active area of research for the sensors community~\cite{cmos_new_relevant_2023, cmos_new_relevant_2025}, we benchmark a combination of logarithmic and learned $\gamma$ quantization. That is, we use $\mathcal{X}_{\text{norm}} = \mathcal{X}_{\mathrm{log}}$ in \cref{eqn:our_gamma_teaser}.
\end{enumerate}

\subsection{Learning Task-specific Low-bit Quantization} 
\label{sec:method:learning}
 We learn the parameter $\gamma$ of \method{} and `Log + \method{}' jointly with the neural network parameters for the specific task by optimizing
 \begin{equation}
 \min_{\theta, \gamma} ~\mathbb{E}_{(\mathcal{{X}},\mathrm{y})}\left[\mathcal{L}(\eta(\mathcal{Q}(\mathcal{{X}}, \gamma)),\mathrm{y};\theta)\right],
     \label{eq:joint_learning}
 \end{equation}
  where $\eta$ is the task-specific neural network with parameters $\theta$, and where $\mathcal{L}$ is a suitable loss function comparing the network output and the ground truth prediction $\mathrm{y}$. 

Since \rawdet{} offers a unique multi-scenario optimization setting, we capitalize, and attempt learning $\gamma$ values for different scenarios.
Using one $\gamma$ value over the entire dataset is denoted as 1-$\gamma$.
When using two $\gamma$ value we denote it as 2-$\gamma$, with the intention of them conditioning to the time of day, \ie one $\gamma$ for daytime and one for nighttime images.
Lastly, we test using five $\gamma$ values, denoted as 5- $\gamma$, allowing one optimized to each sensor.
  
 For optimization, since the quantization operation stops the flow of gradients in the backward pass, we use a straight-through estimator to allow gradient-based optimization of  $\gamma$.   
 Please note that $\gamma<0$ would lead to undesirable behavior, i.e.~, $\mathcal{X}_{\mathrm{Q}} \geq 2^\mathrm{\hat{N}}-1$ when $\mathrm{N} \geq \mathrm{\hat{N}}$ (which is the case when we want to quantize to lower bit depths).
 To avoid this, we clamp $\gamma$ to be non-negative, using $\mathrm{ReLU(.)}$~\cite{relu}.
To take into account Bayer patterned images in the RAW dataset, we extract red, green, and blue channels from the Bayer patterned image and downsample it using nearest neighbor interpolation to simulate the capture of a low-resolution, low-bit-depth image, which is provided as input to the neural network. 
This pipeline for joint training of $\gamma$ and task-specific neural network parameters $\theta$ using quantized RAW images is 
the methodology for the forthcoming evaluations.


\section{Experiments}
\begin{figure}
    \centering
    \includegraphics[width=1.0\linewidth]{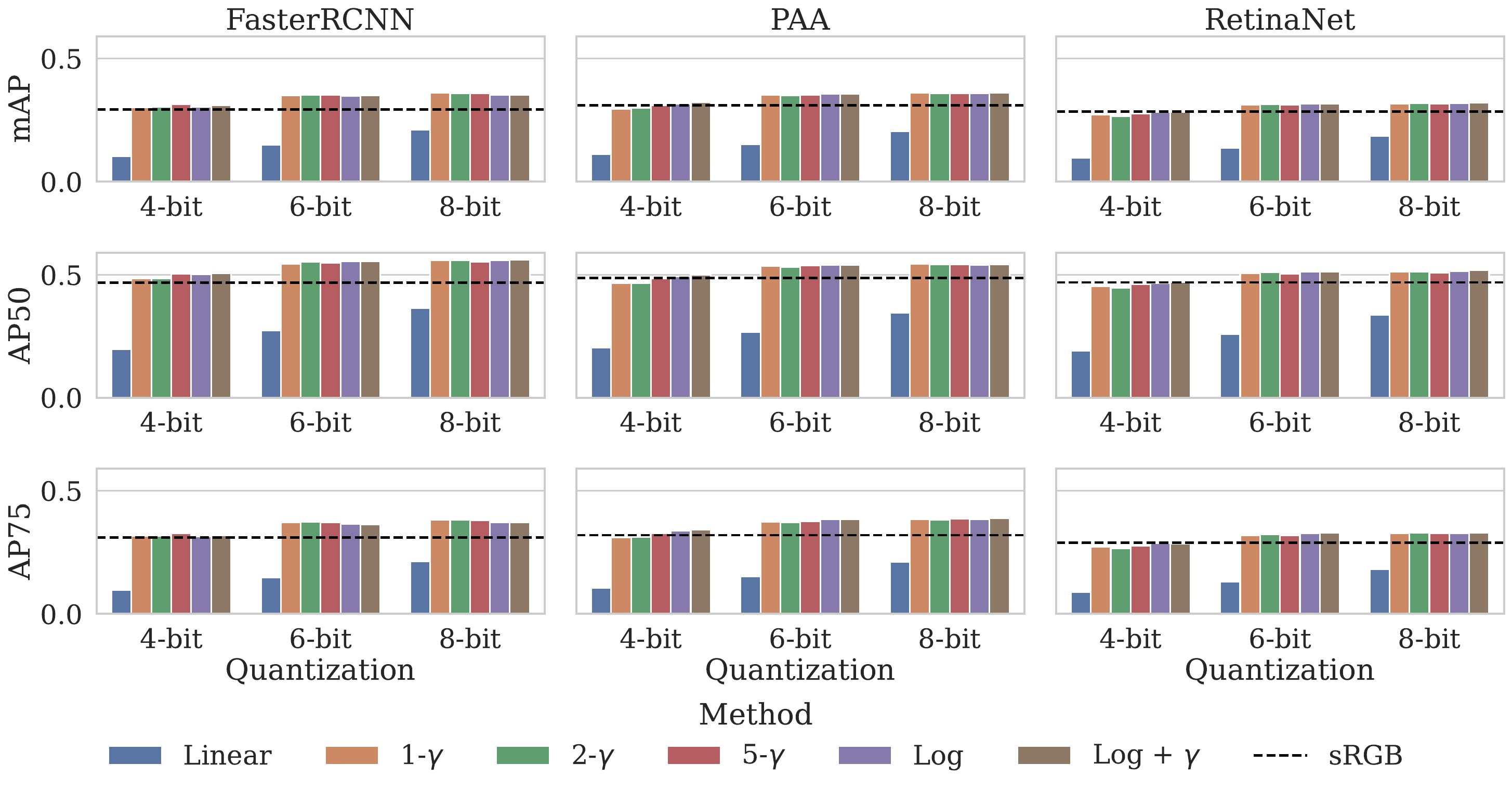}
    \caption{Benchmarking performance on \rawdet{}. Baselines such as logarithmic quantization and jointly learnt $\gamma$ improve results across quantization levels and architectures.   }
    \label{fig:results_raw_det_7}
\end{figure}
For object detection we use the mmdetection~\cite{chen2019mmdetection} framework from openmmlab. 
Following previous works~\cite{xu2023RAODdataset,omid2014pascalraw} that perform object detection on RAW images, we conduct experiments using three architectures  Faster-RCNN~\cite{faster_rcnn}, RetinaNet~\cite{lin2017focal}, and PAA~\cite{kim2020probabilistic_paa} models with a pre-trained ResNet50~\cite{he2016deep} backbone. 
For fair comparison with sRGB counterpart, we extract the `R'(Red), `G' (Green), and `B'(Blue) channels from the Bayer patterned RAW input, average over the two `G' channels and 
report the following quantitative evaluation metrics: mean average precision (mAP) is calculated by averaging the Average Precision (AP) values across the intersection over union (IoU) thresholds ranging from [0.5, 0.95] with a step size of 0.05, resulting in 10 threshold values. 
We also report performance with IoU-thresholds of 0.5 and 0.75 (AP50 and AP75) for each case. 
We use an ImageNet-1k~\cite{imagenet} pretrained backbone and freeze the first stage of the backbone during training. 
We follow the multi-scale training setup commonly used~\cite{faster_rcnn,lin2017focal}, during training, and at test time, the shorter edge length is kept at 800. 
We use a batch size of 16 and train the model for 140 epochs. 
We use SGD with Nestrov Momentum~\cite{sgd_nestrov} as the optimizer with a weight decay of $1e^{-3}$. 
For the PASCAL RAW dataset, we use the same train/test split as proposed by the original paper. For the Nikon, Sony, and Zurich datasets, we do an 80-20 train/test split of the entire dataset. For RAOD, we use the publicly available validation set as the test set. All dataset splits and statistics are provided in \cref{fig:dataset_statistics}, whereas their sensor models are provided in  \cref{tab:sensor_statistic} in the appendix. 

\begin{figure}[t!]
    \centering
    \includegraphics[width=1.0\linewidth]{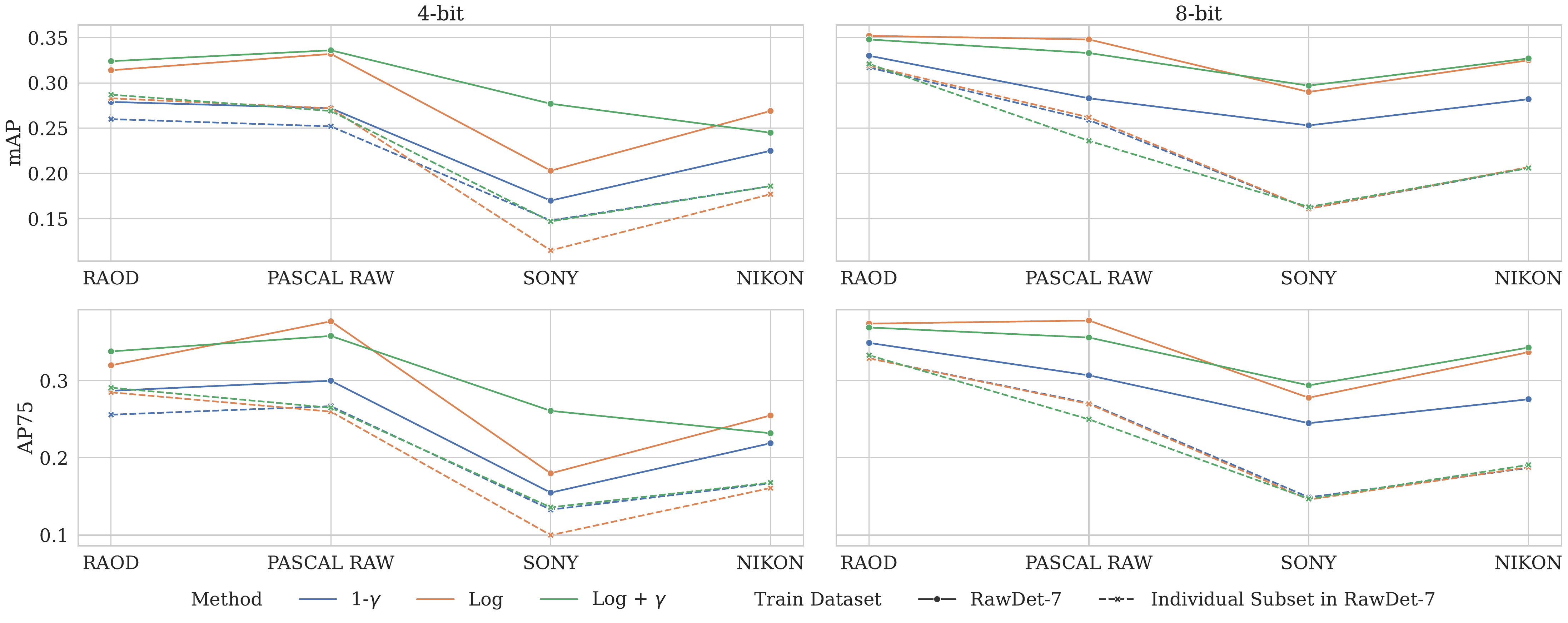}
    \caption{Results for training on combined \rawdet{} when evaluated on each subset, vs.~training only on a subset and evaluated the subset's test set. Combining improves results on all subsets. The model used is Faster RCNN. }
    \label{fig:combined_individual_training}
    \vspace{-0.5em}
\end{figure}

\subsection{Benchmarking Performance on \rawdet}
We report the performance of various object detection networks in the proposed \rawdet{} benchmark, for different quantization levels in \cref{fig:results_raw_det_7}. 
Here we observe that na\"ively reducing the bit depth through linear scaling and quantization to a low bit depth degrades the quality of fine-grained details that are critical for tasks like object detection. 
In comparison, the non-linear scaling and quantization methods perform significantly better at all the bit depths. 
Optimizing only a single additional parameter $\gamma$ (for  $\gamma$ -scaling) jointly with the downstream network during training, we observe a significant improvement in detection performance over using both linearly scaled and quantized inputs, and sRGB inputs. 
This shows that even a lightweight, task-aware mapping of the input for quantization can lead to meaningful improvements over na\"ive quantization pipelines. 
Logarithmic scaling and quantization, albeit fixed, performs on par with or better than learned $\gamma$ scaling. 
Among the different target bit depths considered, we observe that 4-bit linear scaling and quantization performs worse across all architectures. 
Improvements are observed by applying a logarithm or the learned $\gamma$ scaling before the quantization. 
The improvements are on par with the sRGB setting, indicating that a simple baseline (such as log or learned  $\gamma$ ), when applied before the quantization, alleviates the need for an ISP. We also learn separate $\gamma$ during training for day and night images as well as separate $\gamma$ for different sensors indicated by 2-$\gamma$  and 5-$\gamma$, respectively in \cref{fig:results_raw_det_7}. 
We observe a marginal difference in performance while training with multiple $\gamma$ values, while combining logarithmic scaling with learned $\gamma$ scaling provides some gain in detection performance when compared to solely using logarithmic scaling for networks trained from scratch.
 
\subsection{Added Value by the Contributed Annotations}
\begin{figure}
    \centering
    \includegraphics[width=1.0\linewidth,clip]{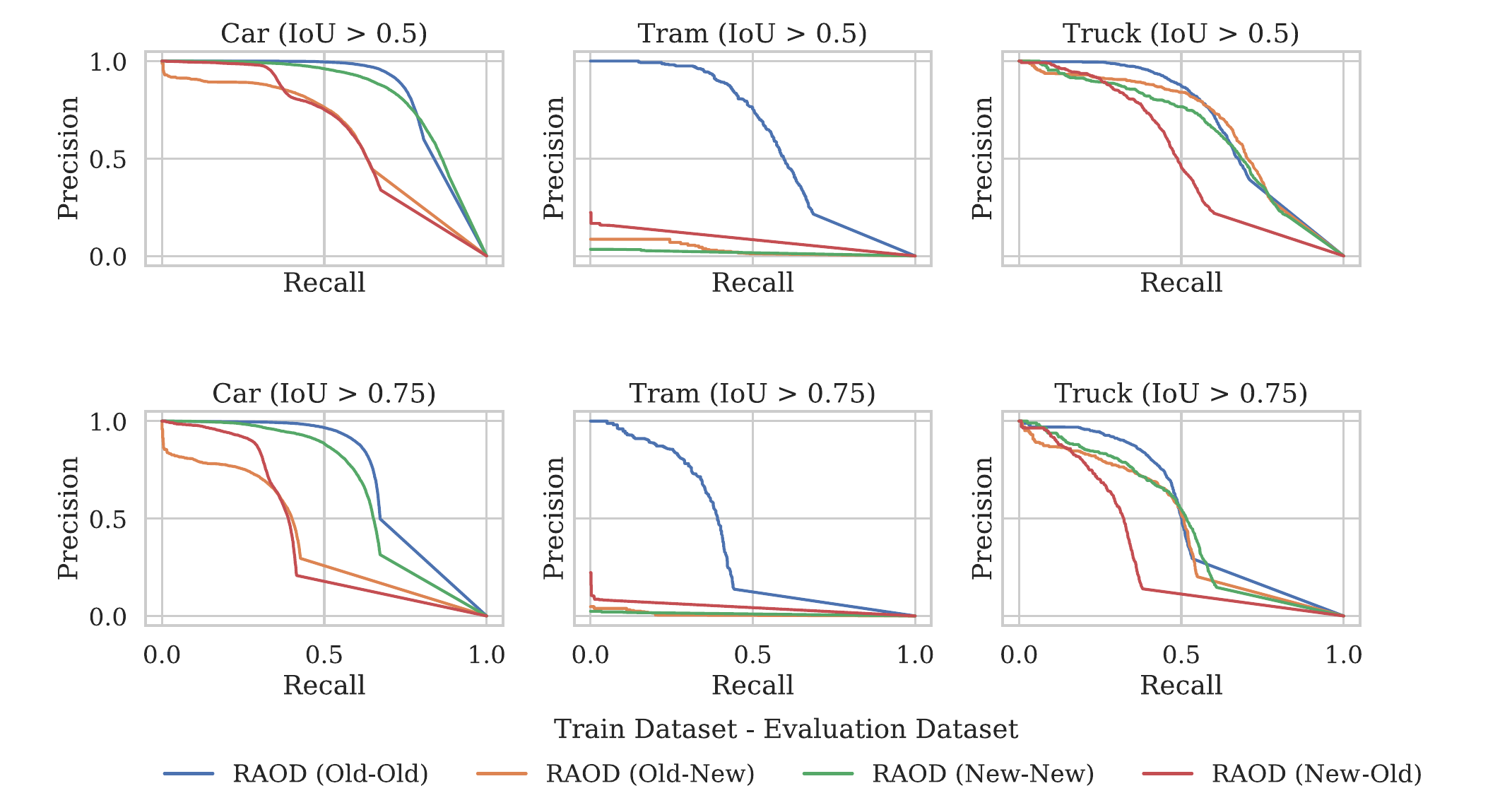}
    \caption{Precision Recall Curves for RAOD. ``Old'' refers to the annotations as proposed in the original dataset, whereas ``New'' means the annotations in \rawdet{}. All the models are trained on FasterRCNN, for 8-bit quantization and jointly learnt with 1 gamma.
    As mentioned in the legend, the keys are ``(Train Dataset - Evaluation Dataset)''.}
    \label{fig:pr_raod}
\end{figure}
\begin{table}[t]
\centering
\caption{Experiments with large VLM, MM-Grounding-DINO~\cite{mm_grounding_dino} with a Swin-T~\cite{liu2021Swin} backbone. For experiments with \method{} and Log + \method{} we finetune only the \method{} parameter while keeping the VLM frozen, for the other methods, we evaluate zero-shot.}
\scriptsize
\begin{tabular}{lccccc}
\toprule
Quantization & Method & mAP & AP50 & AP75 & AP95 \\ 
\midrule
RGB & sRGB & 0.188 & 0.215 & 0.204 & 0.088 \\
\midrule
\multirow{4}{*}{4-bit RAW} & Linear & 0.047 & 0.060 & 0.053 & 0.008 \\
 & Log & 0.114 & 0.138 & 0.127 & 0.034 \\
 & \method{} & 0.104 & 0.125 & 0.115 & 0.035 \\
 & Log + \method{} & 0.135 & 0.161 & 0.149 & 0.048 \\
\midrule
\multirow{4}{*}{6-bit RAW} & Linear & 0.066 & 0.081 & 0.074 & 0.014 \\
 & Log & 0.149 & 0.182 & 0.164 & 0.042 \\
 & \method{} & 0.177 & 0.206 & 0.193 & 0.072 \\
 & Log + \method{} & 0.198 & 0.230 & 0.217 & 0.085 \\
\midrule
\multirow{4}{*}{8-bit RAW}& Linear & 0.074 & 0.091 & 0.083 & 0.017 \\
 & Log & 0.151 & 0.186 & 0.167 & 0.041 \\
 & \method{} & 0.177 & 0.207 & 0.193 & 0.070 \\
 & Log + \method{} & 0.202 & 0.233 & 0.220 & 0.090 \\
\midrule
\multirow{4}{*}{12-bit RAW} & Linear & 0.075 & 0.092 & 0.084 & 0.017 \\
 & Log & 0.152 & 0.187 & 0.168 & 0.041 \\
 & \method{} & 0.167 & 0.196 & 0.183 & 0.066 \\
 & Log + \method{} & 0.201 & 0.232 & 0.219 & 0.089 \\
\bottomrule
\end{tabular}
\label{tab:mm_grounding_dino}
\vspace{-1em}
\end{table}
\cref{fig:pr_raod} shows the precision-recall curves for the RAOD dataset in four different settings. ``Old-Old'' means that the model was trained with original annotations and tested with them as well, whereas ``Old-New'' indicates a different test setting, testing with newly proposed annotations, and likewise for ``New-New'' and ``New-Old''.
Since the annotations in both cases are slightly different, we only compute the precision-recall curves for 4 mutually common annotations \ie \textit{`Car'}, \textit{`Tram'}, \textit{`Truck'}, and \textit{`Person'}. 
We observe that the precision when tested with new annotations is lower for the same recall value. 
This indicates that the difficulty level of these annotations is higher compared to the original dataset. \cref{fig:combined_individual_training} shows that training on the proposed \rawdet{} dataset improves performance across individual datasets, compared to training the model individually with every subset. 
This indicates that combining multi-bit depth raw images during training helps the model generalize well on every subset of data, which further proves the usefulness of \rawdet{}.

\begin{figure}[!t]
    \centering
    \resizebox{\linewidth}{!}{
    \begin{tabular}{@{}c@{\hspace{0.5mm}}c@{\hspace{0.5mm}}c@{\hspace{0.5mm}}c@{\hspace{0.5mm}}c@{\hspace{0.5mm}}c@{}}
         \textbf{\Large Linear} & \textbf{\Large Log} & {\Large \method{}} & {\Large \textbf{Log} + \method{}} & \textbf{\Large sRGB} & \textbf{\Large Ground Truth} \\
         \includegraphics[width=3.75cm,height=2.85cm]{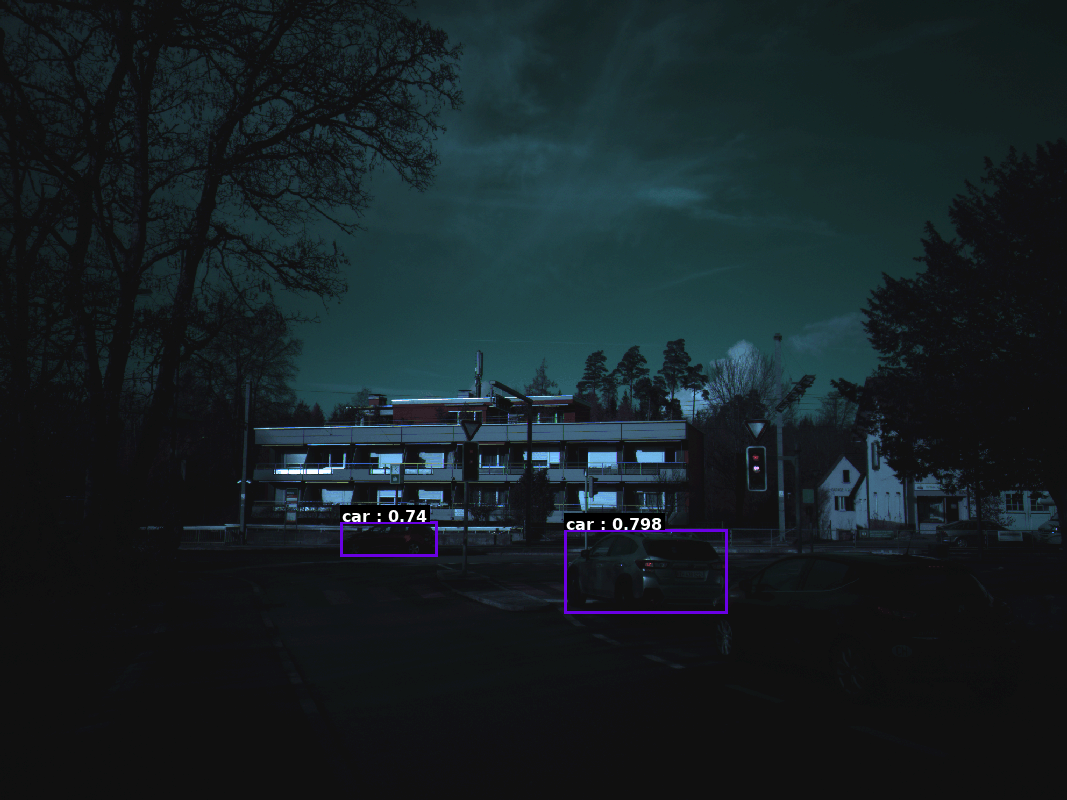} &
         \includegraphics[width=3.75cm,height=2.85cm]{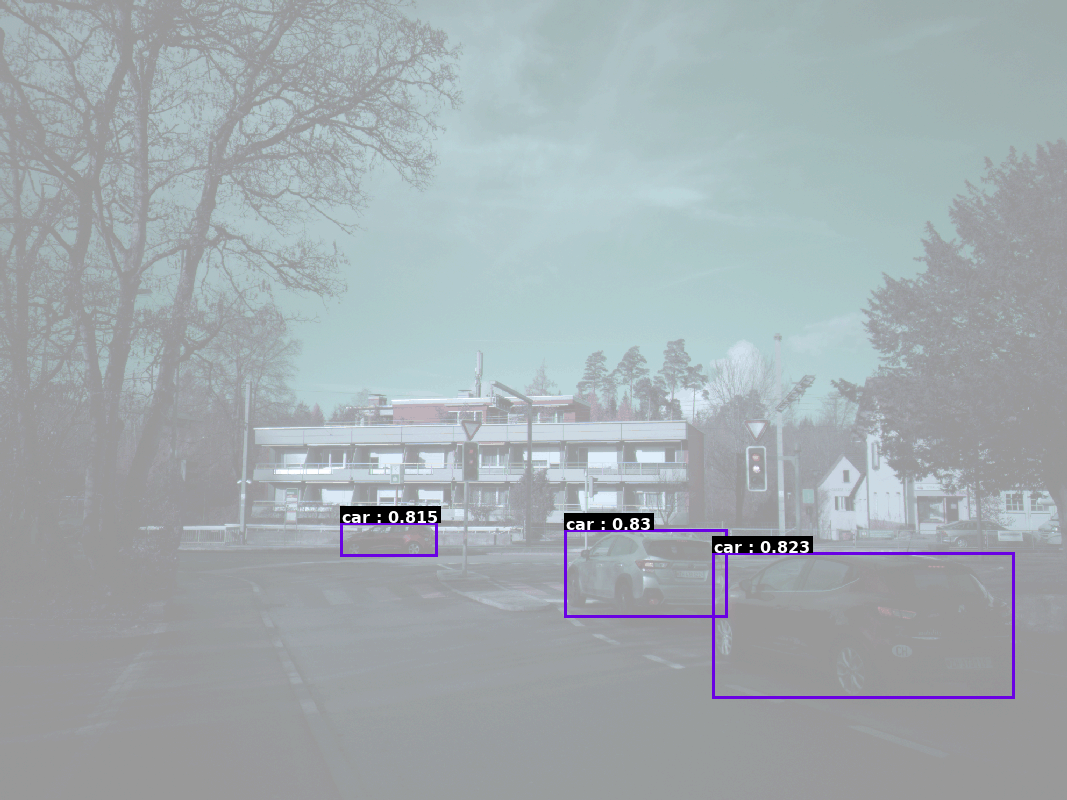} &
         \includegraphics[width=3.75cm,height=2.85cm]{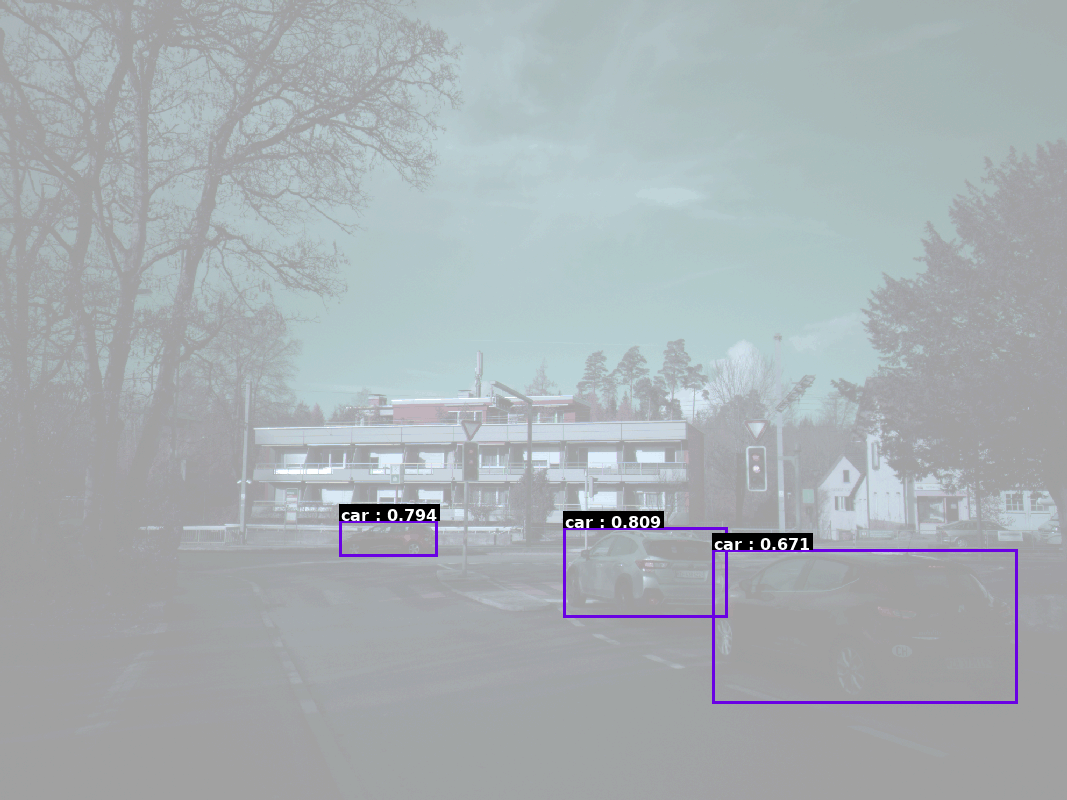} &
         \includegraphics[width=3.75cm,height=2.85cm]{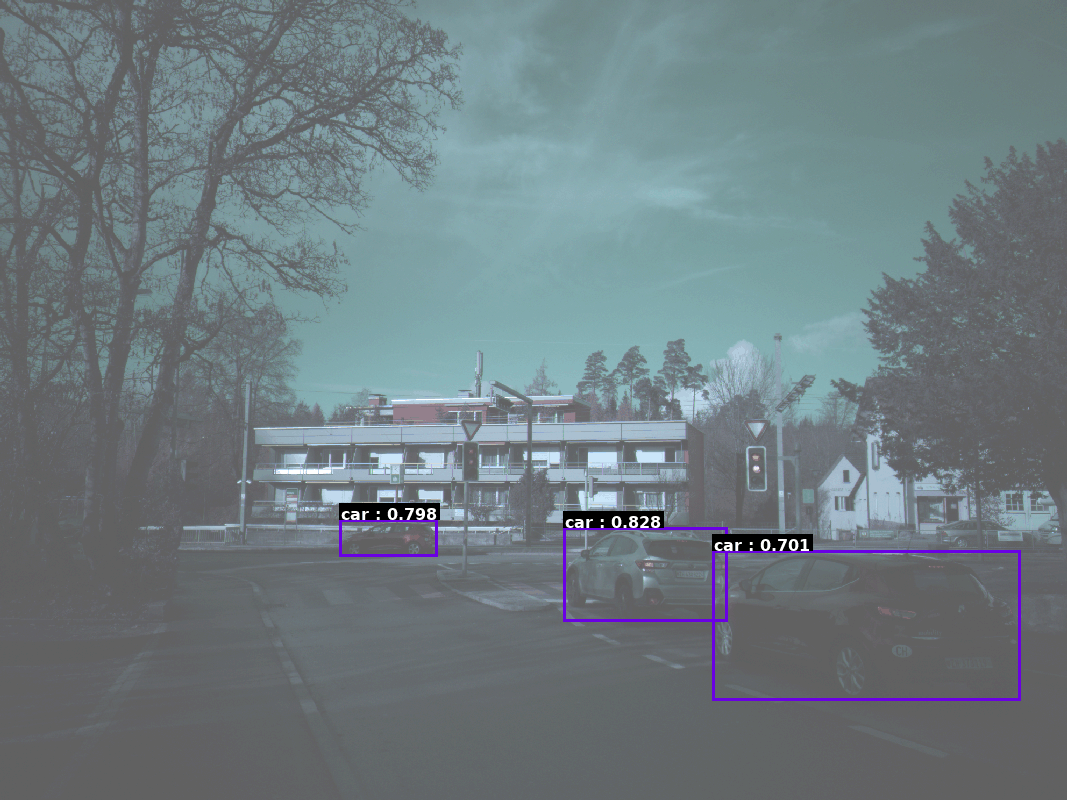} &
         \includegraphics[width=3.75cm,height=2.85cm]{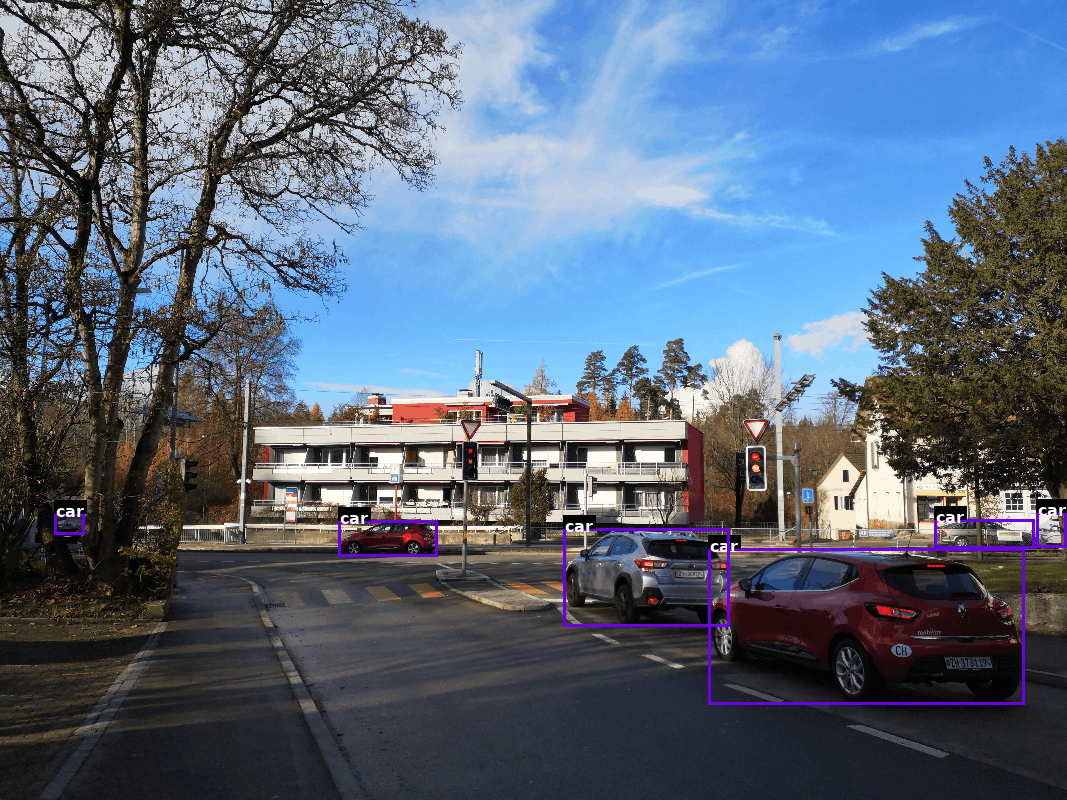} &
         \includegraphics[width=3.75cm,height=2.85cm]{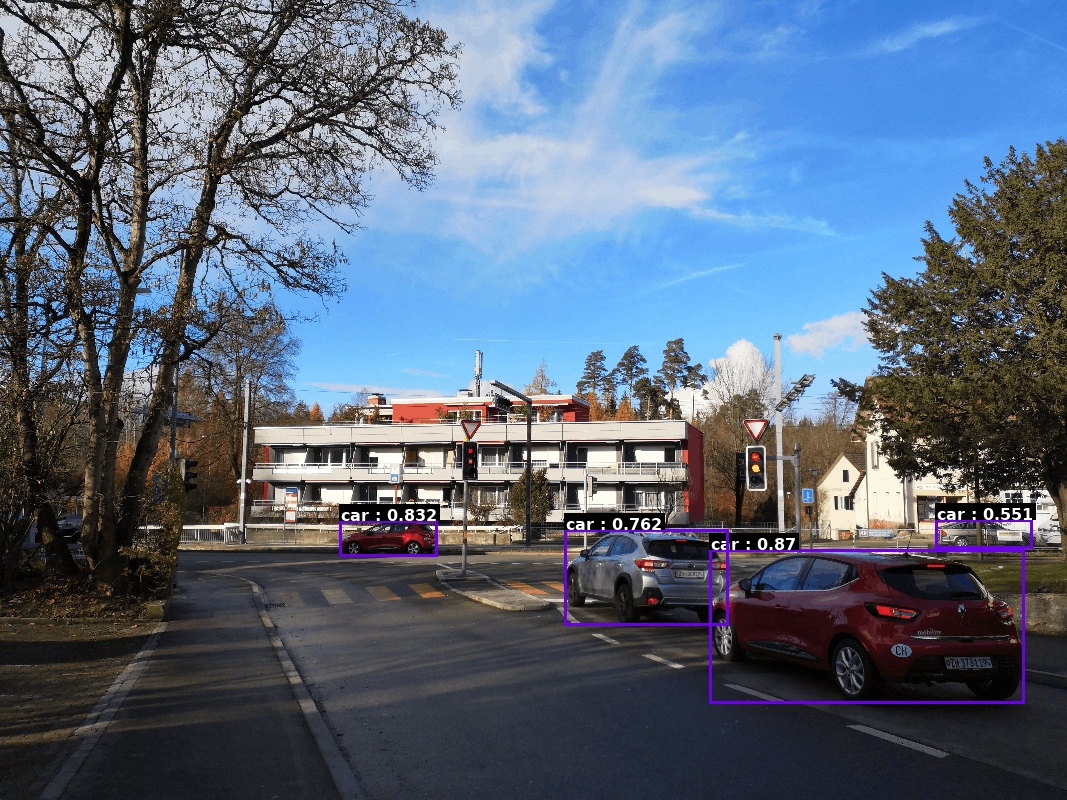} \\

         \includegraphics[width=3.75cm,height=2.85cm]{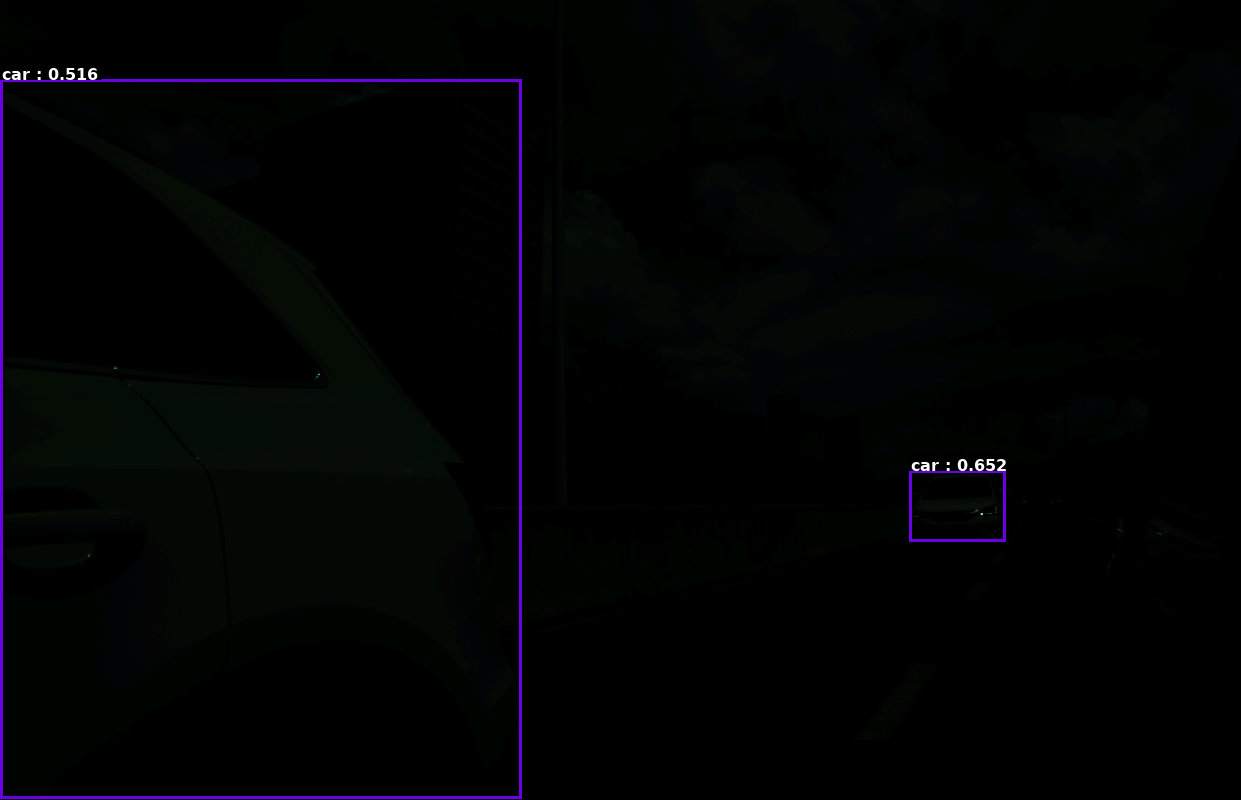} &
         \includegraphics[width=3.75cm,height=2.85cm]{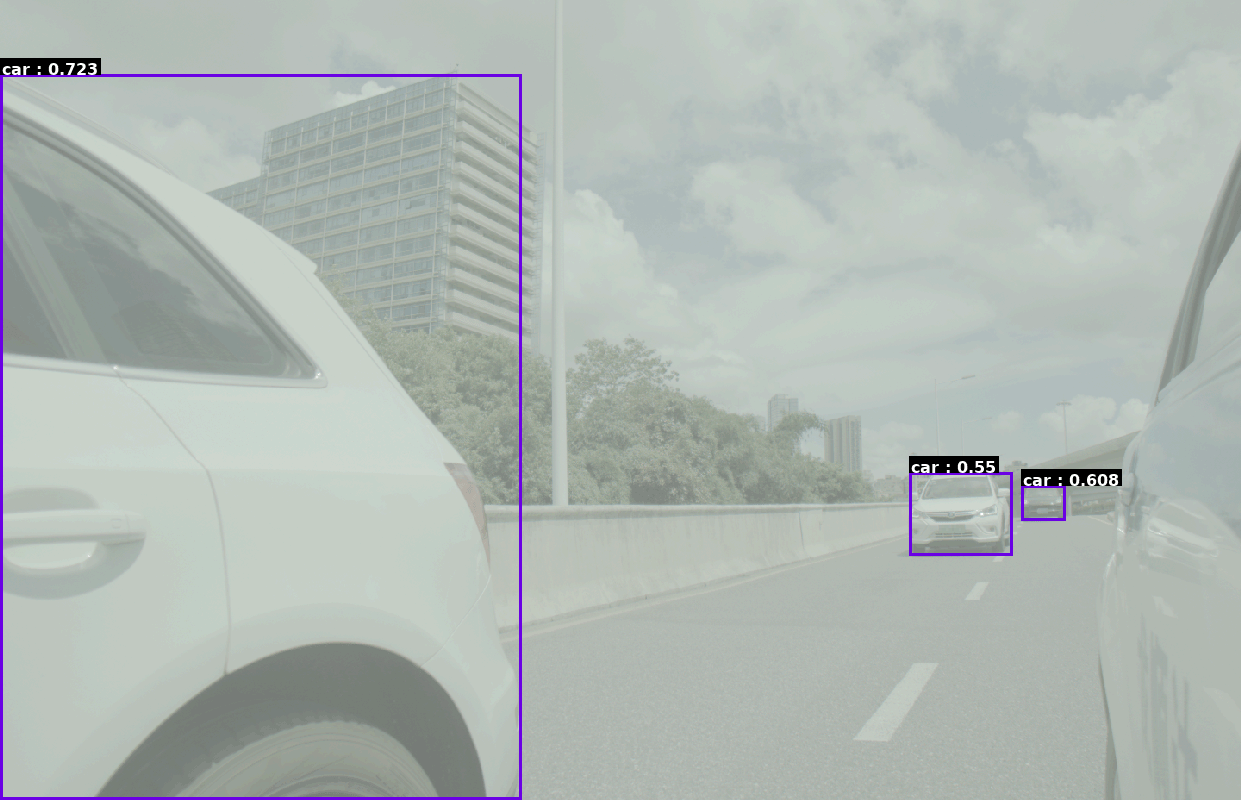} &
         \includegraphics[width=3.75cm,height=2.85cm]{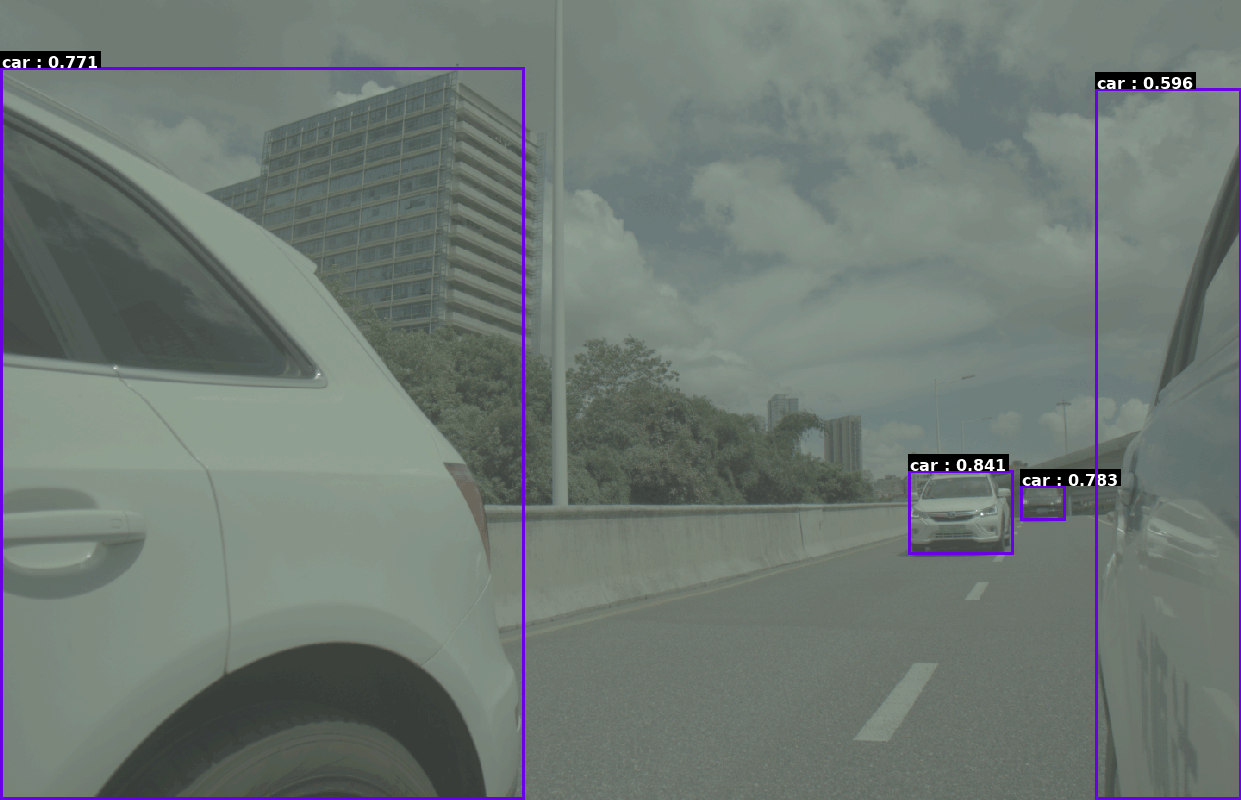} &
         \includegraphics[width=3.75cm,height=2.85cm]{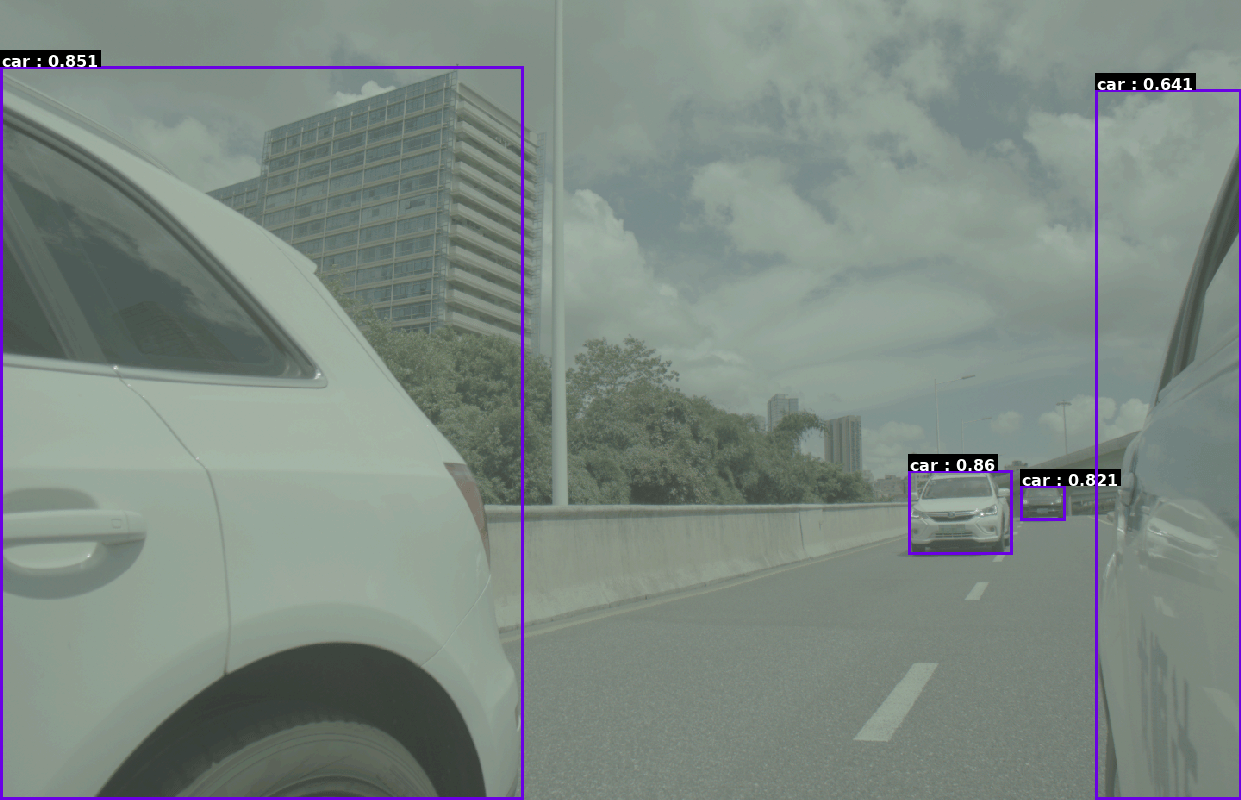} &
         \includegraphics[width=3.75cm,height=2.85cm]{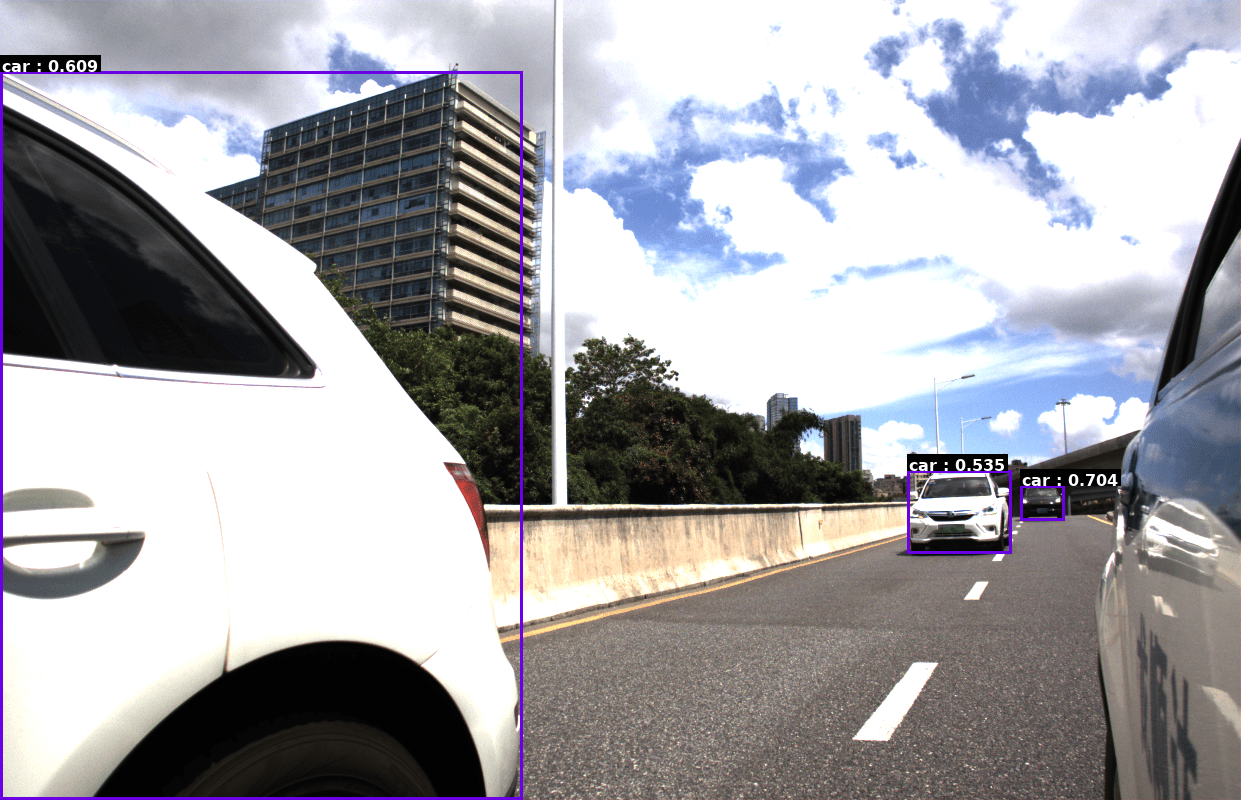} &
         \includegraphics[width=3.75cm,height=2.85cm]{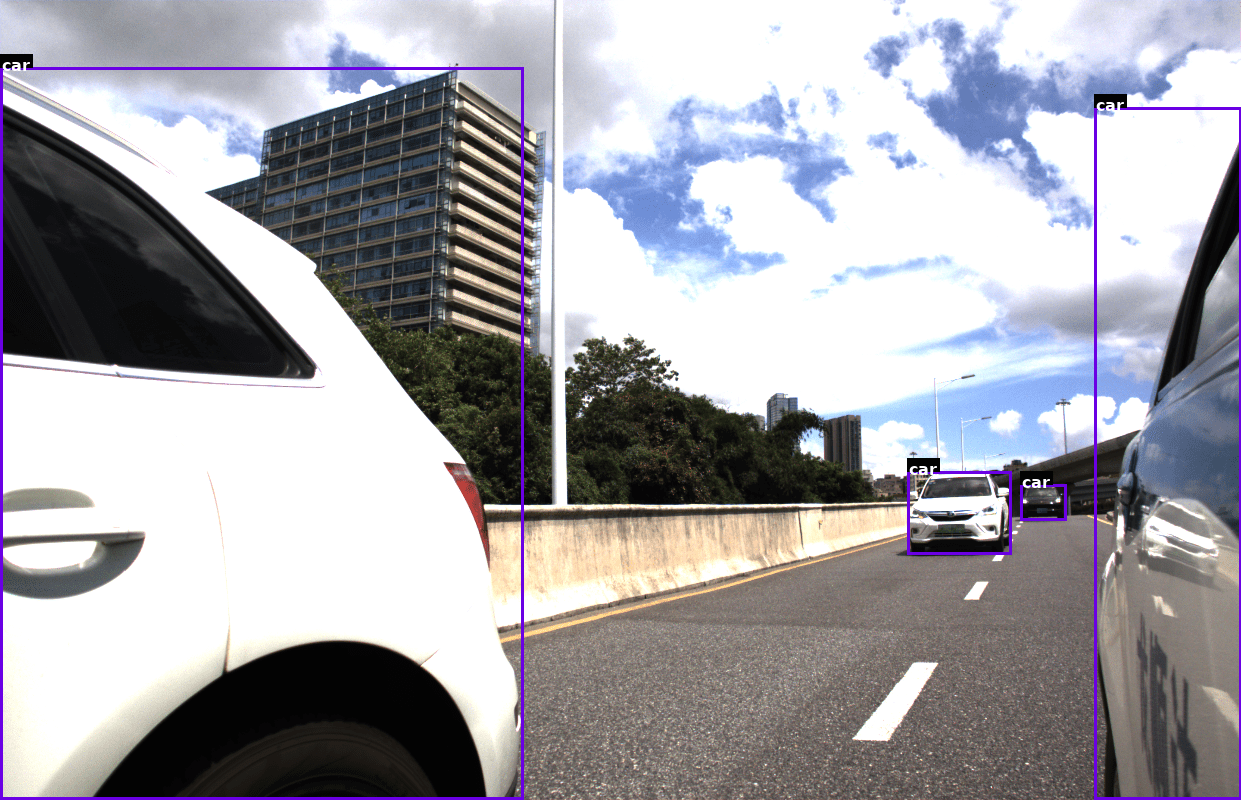} \\

         \includegraphics[width=3.75cm,height=2.85cm]{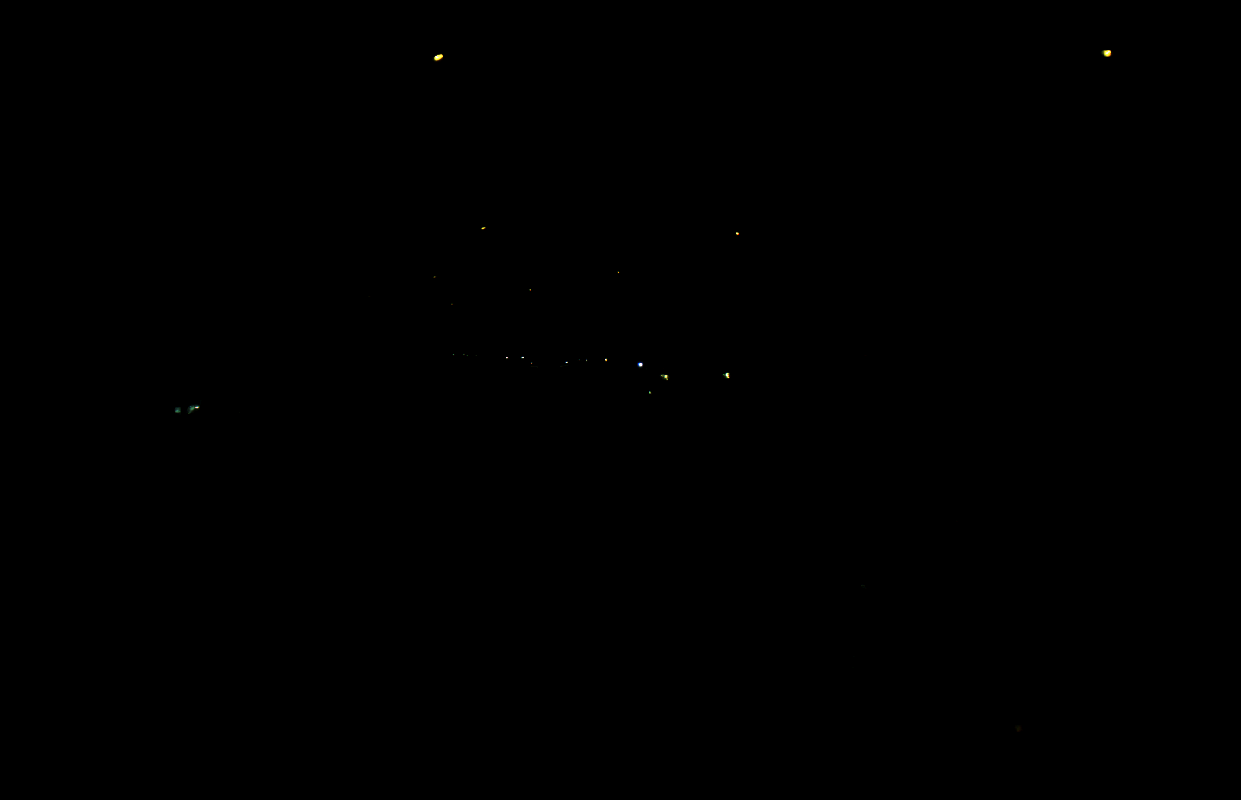} &
         \includegraphics[width=3.75cm,height=2.85cm]{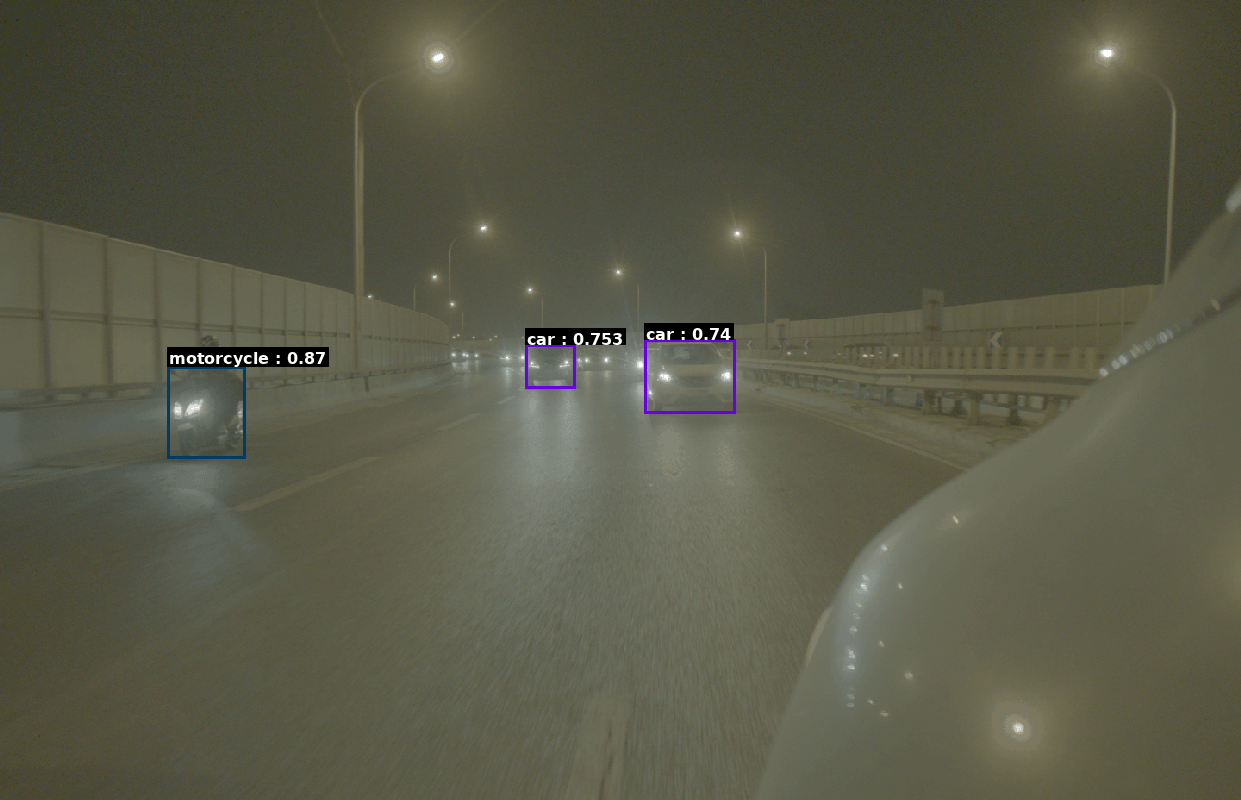} &
         \includegraphics[width=3.75cm,height=2.85cm]{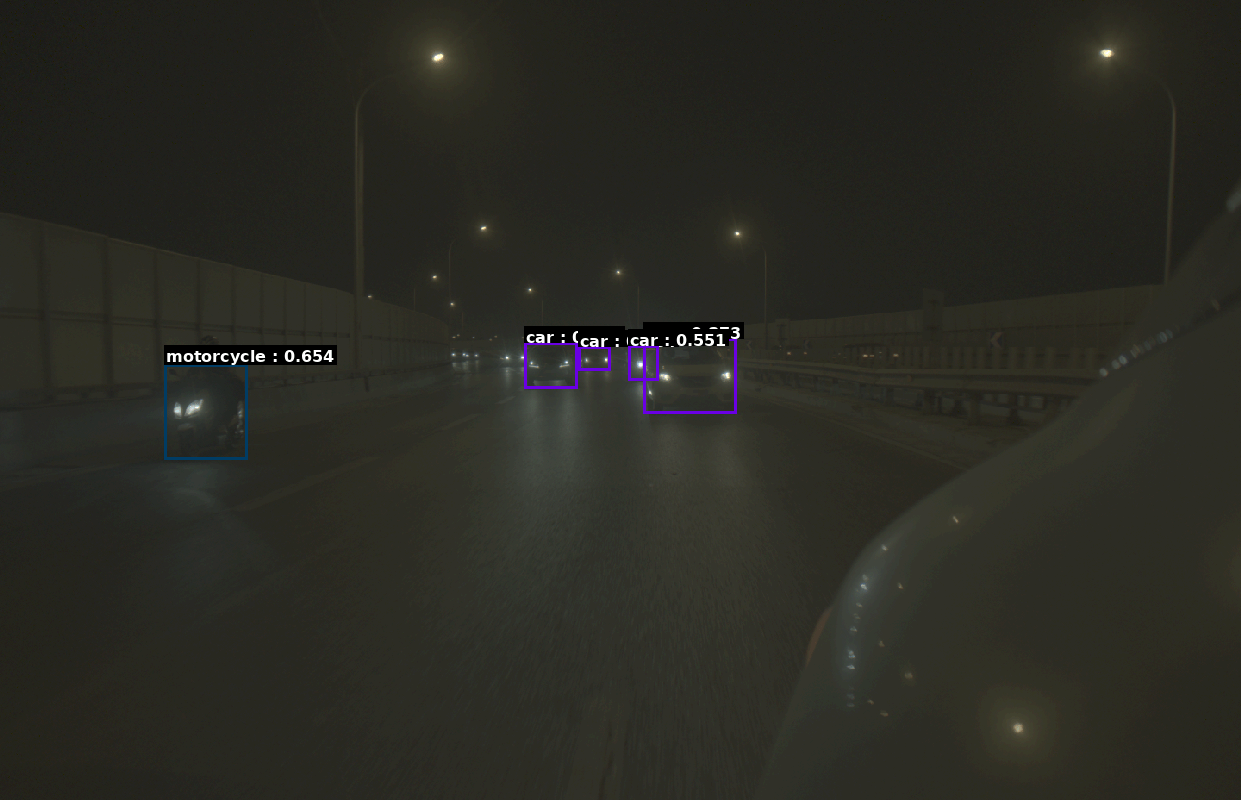} &
         \includegraphics[width=3.75cm,height=2.85cm]{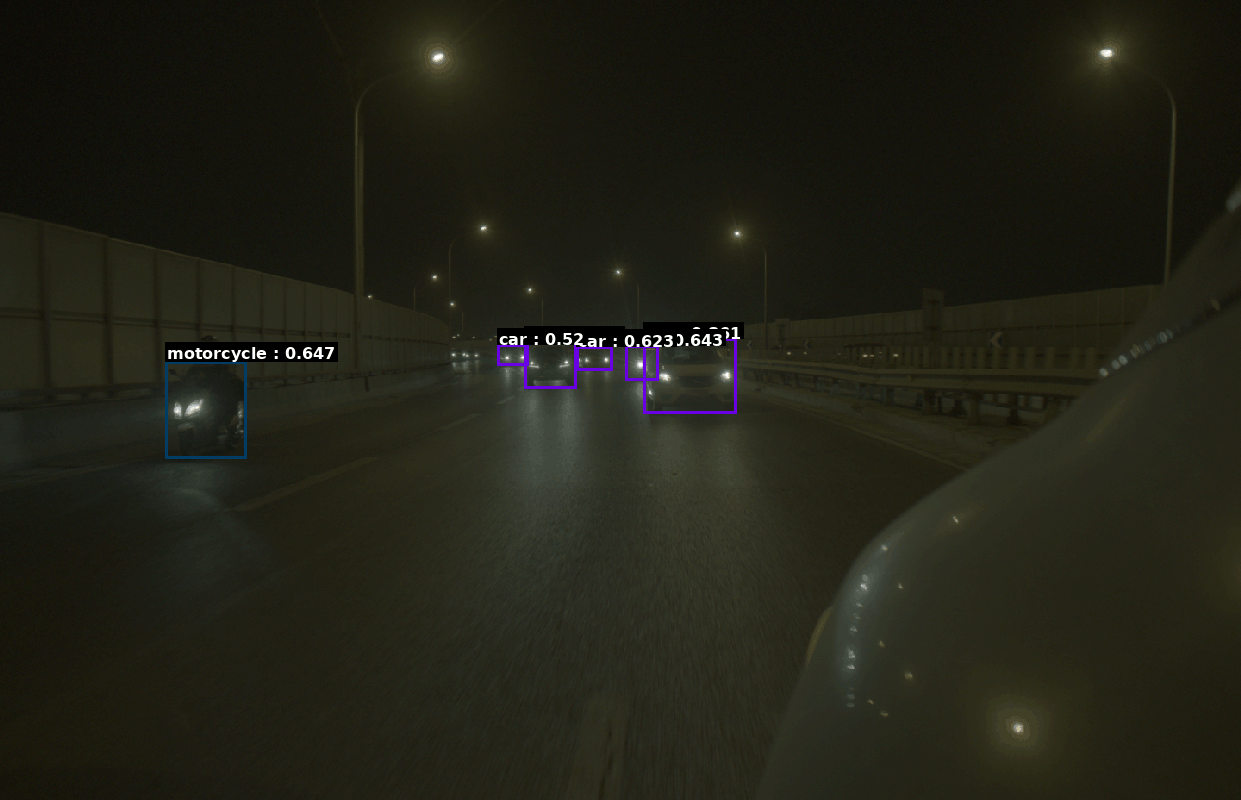} &
         \includegraphics[width=3.75cm,height=2.85cm]{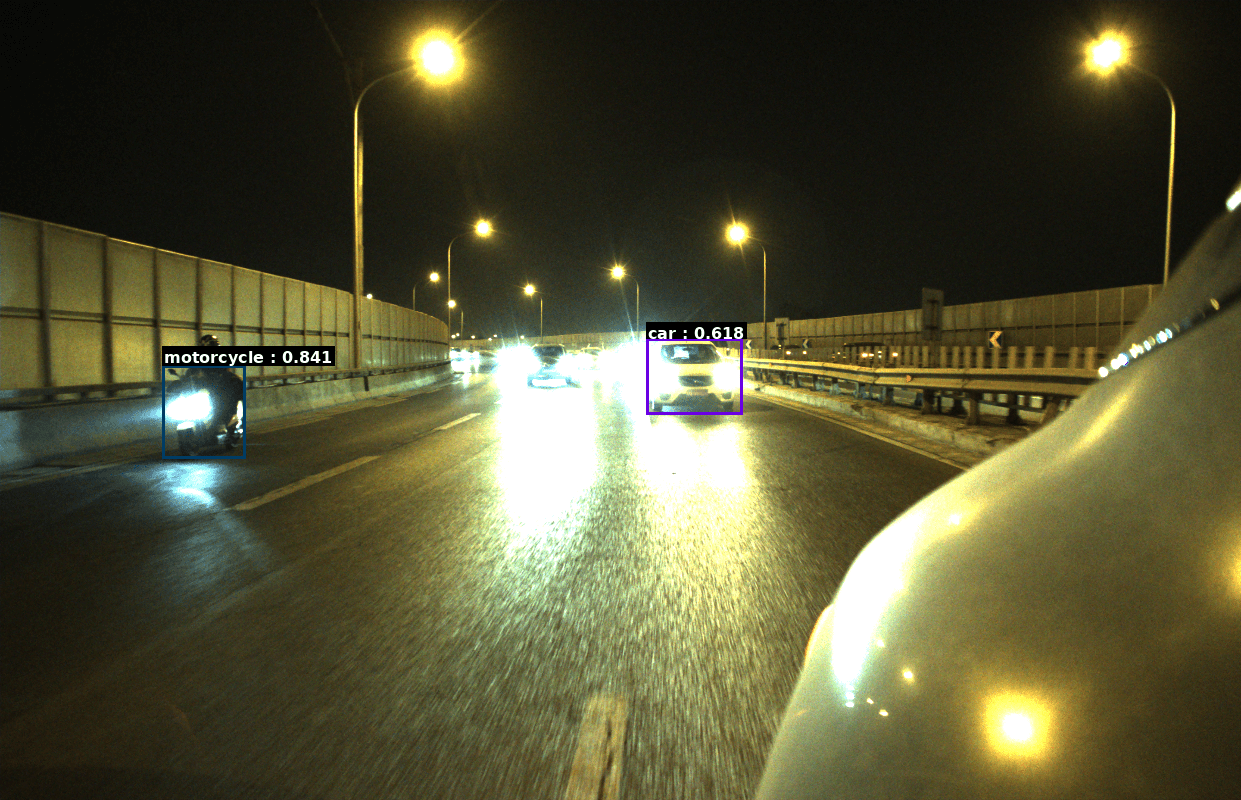} &
         \includegraphics[width=3.75cm,height=2.85cm]{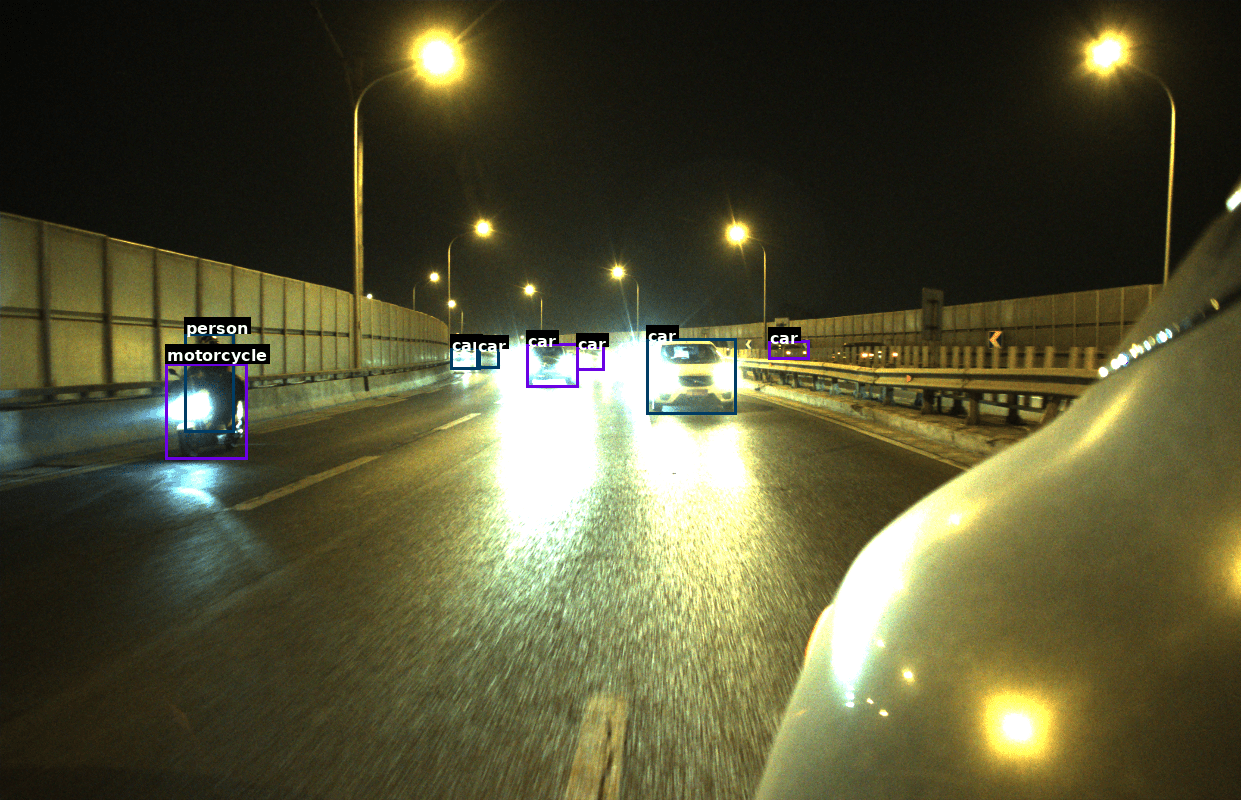} \\
    \end{tabular}
    }
    \caption{Visualizing predictions made by MM-Grounding-DINO on \rawdet{} at 6-bit quantization, using different methods for predictions on sRGB images and the expected Ground Truth predictions. Her,e for methods $\gamma$ and Log + $\gamma$, we finetune only the $\gamma$ parameter while keeping the VLM frozen; for the other methods, we evaluate zero-shot. We randomly sample three distinct scenarios to demonstrate the versatility of the proposed \rawdet{}.}
    \label{fig:mm_grounding_dino_pred}
    \vspace{-0.5em}
\end{figure}


\subsection{Benchmarking Large VLMs (LVLMs)}
\label{sec:experiments:large_vlms}
\begin{figure}
    \centering
    \includegraphics[width=1.0\linewidth,clip]{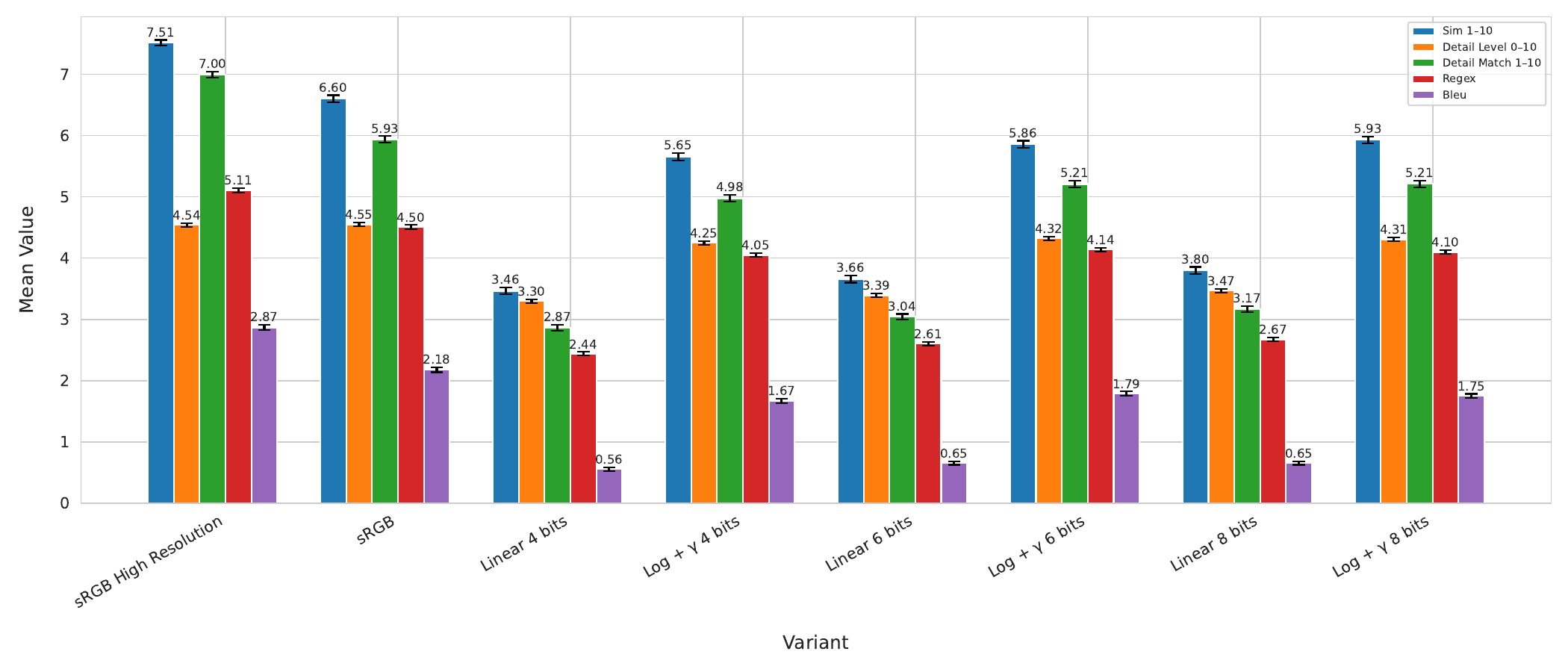}
\caption{Object Description quality across different image variants. Descriptions are generated from linear and log+$\gamma$ variants of quantized images using Gemini, with reference captions obtained from higher-resolution sRGB images. Here, sRGB refers to the sRGB images downsampled 2$\times$ to match the resolution of the RAW images after extracting R, G, and B channels. We multiply BleU and Regex scores by 10 to align them with the other metrics.  Black lines indicate standard errors. The results show that captions generated from processed RAW images (log+$\gamma$) achieve competitive quality, closely matching the reference captions.}
    \label{fig:reasoning}
    \vspace{-1em}
\end{figure}

We also benchmark \rawdet{} on MM-Grounding-DINO for 4 quantization levels and various signal scaling techniques. 
Unfortunately, training or fine-tuning the VLM is computationally infeasible given our limited resources.
Thus, we perform zero-shot evaluations while keeping the VLM frozen.
However, scaling methods with $\gamma$ require some training.
Thus, for these, we keep the VLM frozen and only train the $\gamma$ value with lr 1e-5 and a maximum training budget of 50 epochs.
We observe that Log +\method{} performs the best, even exceeding the performance on sRGB for some quantization levels. 
Log + \method{} highlights the details of the images in a better way as compared to other image-enhancing baselines.
These results are promising, as they show that at multiple-quantization levels, even models trained on a very large amount of sRGB data is able to extract more information out from RAW image input than sRGB input, leading to better performance on RAW input after quantization level over 4-bit depth.
Since at 6-bit depth itself RAW input to MM-Grounding-DINO outperforms sRGB input, we visualize qualitative results for 6-bit depth quantization in \cref{fig:mm_grounding_dino_pred}.

\subsection{Object Description}
\begin{figure*}
    \centering
    \renewcommand{\arraystretch}{1.05}
    \resizebox{0.9\linewidth}{!}{
    \scriptsize
    \begin{tabular}{@{}cp{0.3\linewidth}cp{0.3\linewidth}@{}}
        \midrule
        \multicolumn{2}{l}{\textbf{sRGB:} Similarity  $8.5\pm0.58$ \quad Detail Level $5.0\pm0.82$ \quad Detail Match $8.0\pm0.82$} &\multicolumn{2}{l}{\textbf{sRGB:} Similarity  8.5 $\pm$ 0.71\quad Detail Level  4.5 $\pm$ 2.12 \quad Detail Match  7.5 $\pm$ 0.71 }\\
        \midrule
        \raisebox{-0.95\totalheight}{\includegraphics[width=0.3\linewidth]{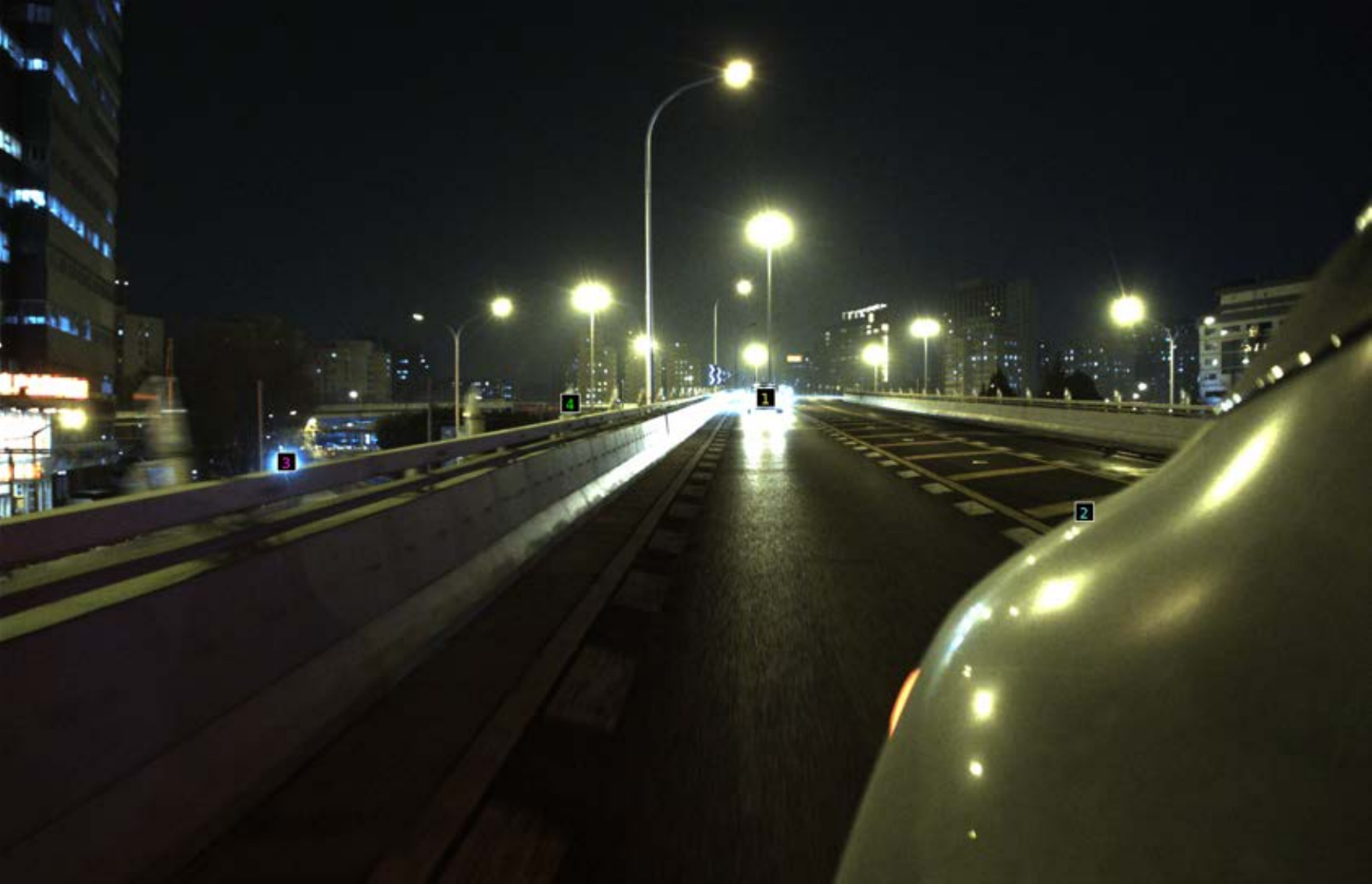}} &
         1. A white car is seen from behind, driving in the left lane of the wet road with its rear lights on.\newline 2. The front right quarter panel of a light-colored metallic car is in the bottom right foreground, with reflections of streetlights visible on its surface.\newline 3.A light-colored car is visible as a motion-blurred shape with a red light streak, located on a lower-level road to the left.\newline 4. A person is visible as a dark silhouette standing on the elevated walkway to the left, behind the road barrier.
        &
\raisebox{-0.95\totalheight}{\includegraphics[width=0.3\linewidth]{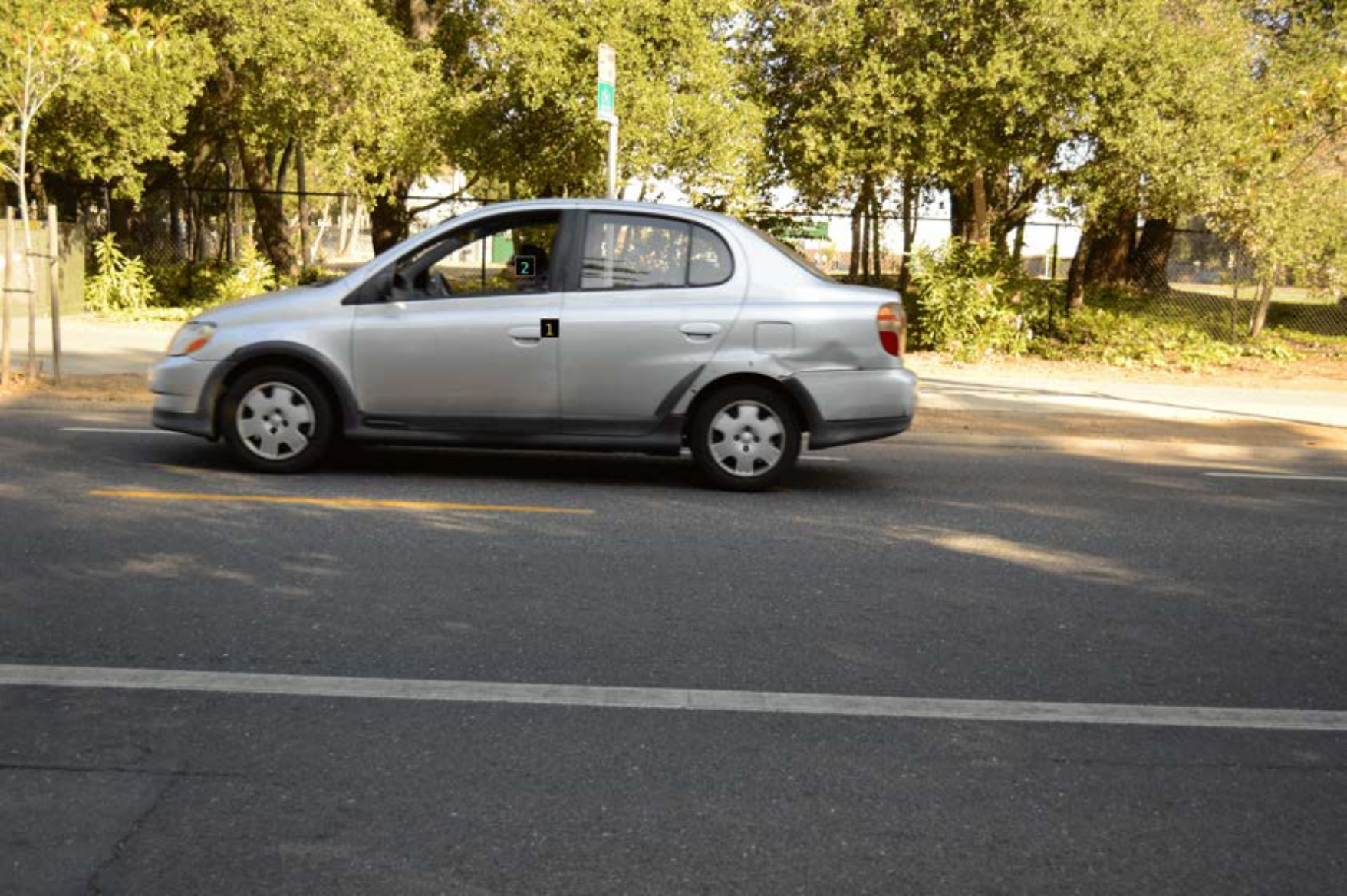}}&
1. A silver sedan is shown from the side, featuring black trim along the bottom and a visible dent on the rear quarter panel. \newline 2. A person with dark hair is seen in silhouette through the driver's side window of the car.\\\midrule
\multicolumn{2}{l}{\textbf{Linear 4-bits:} Similarity $1.0\pm0.00$  \quad   Detail Level $0.0\pm0.00$  \quad  Detail Match $1.0\pm0.00$} &\multicolumn{2}{l}{\textbf{Linear 4-bits:} Similarity 7.5 $\pm$ 2.12 \quad   Detail Level 2.5 $\pm$ 0.71 \quad   Detail Match 5.5 $\pm$ 4.95} \\\midrule
 \raisebox{-0.95\totalheight}{ \includegraphics[width=0.3\linewidth]{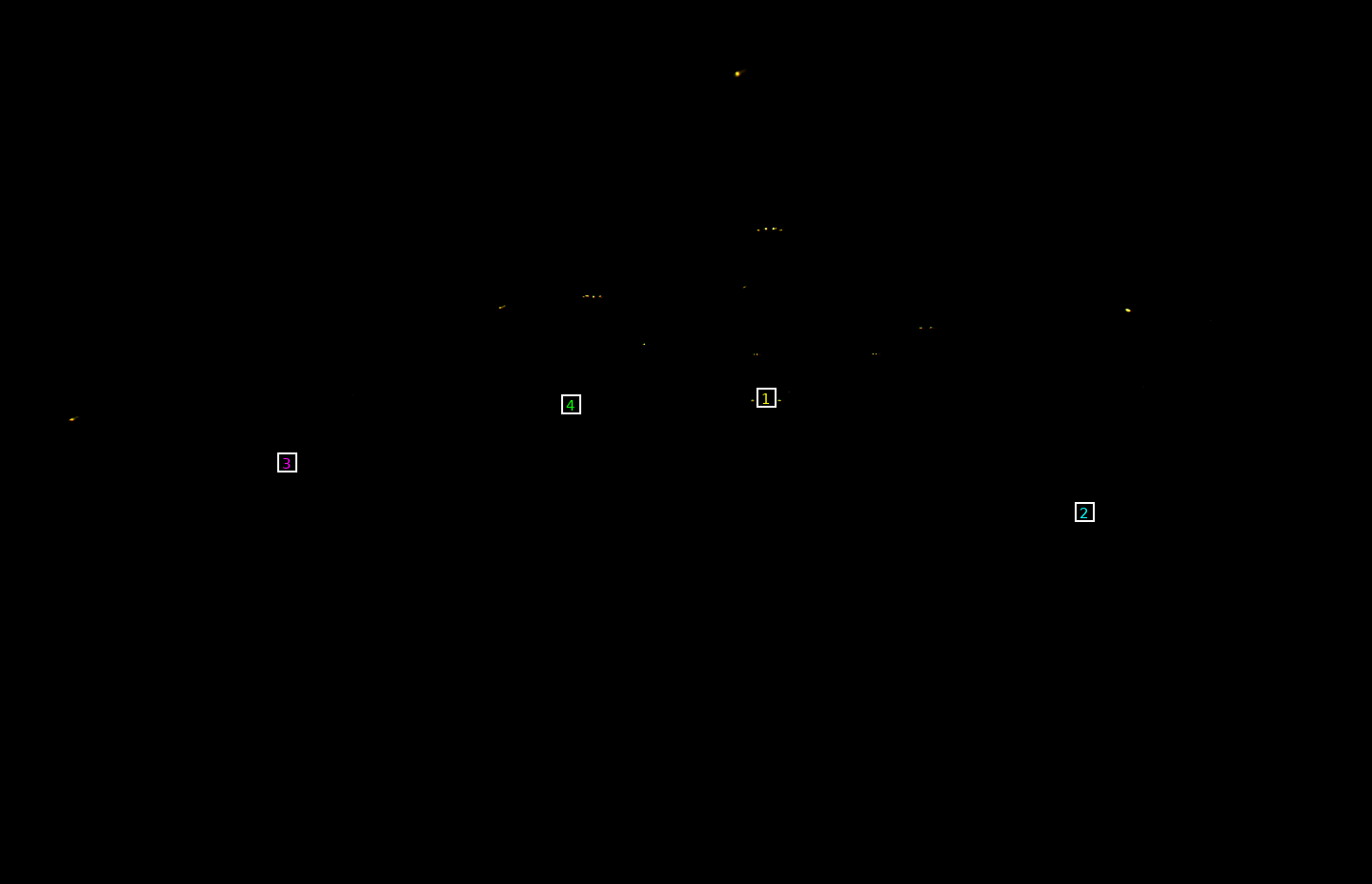}}&1. I cannot see the object\newline 2. I cannot see the object\newline 3. I cannot see the object\newline 4. I cannot see the object
        &   
\raisebox{-0.95\totalheight}{\includegraphics[width=0.3\linewidth]{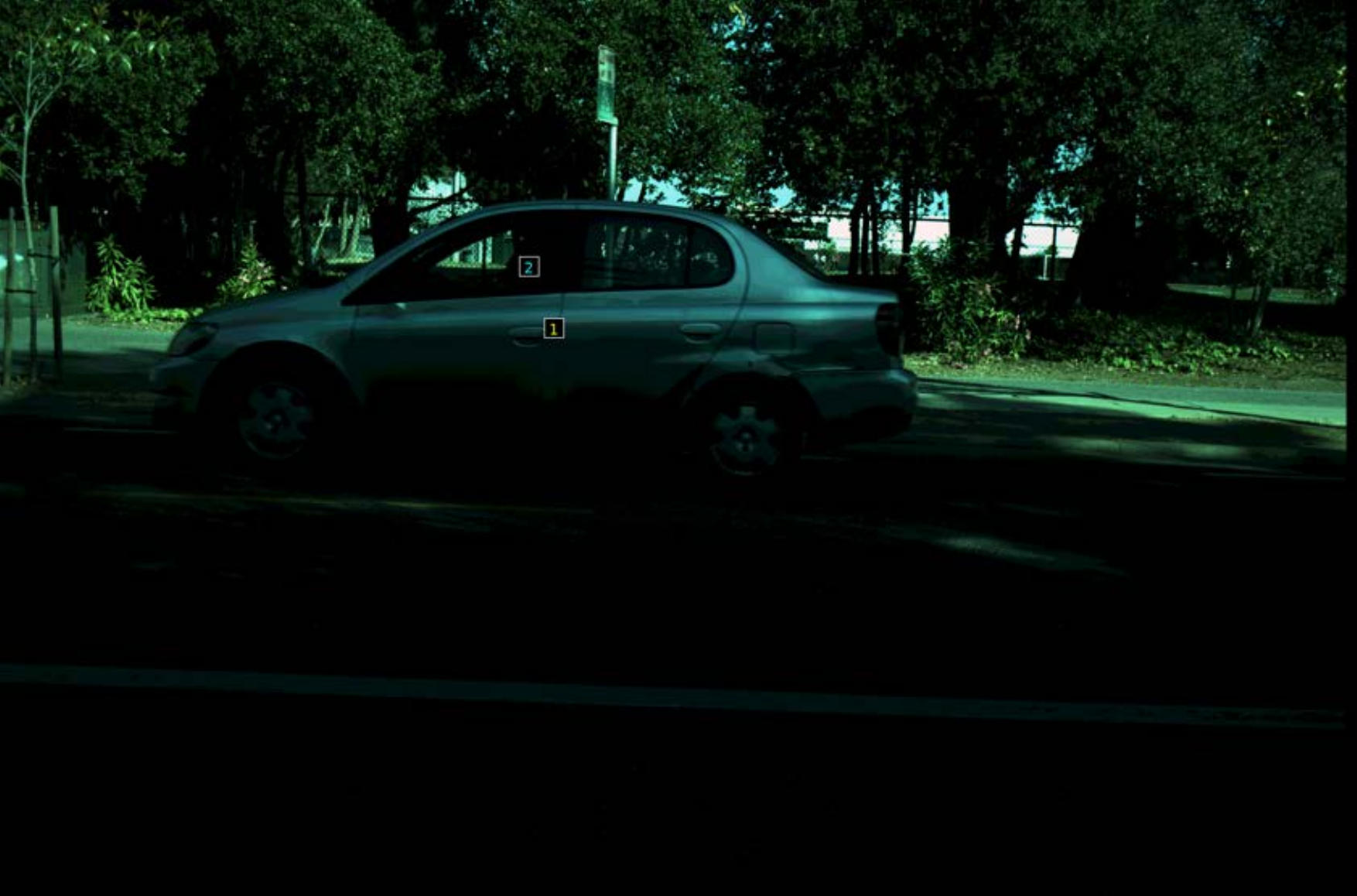}}&
1. A light-colored sedan is parked on the side of a road in shadow. \newline 2. The silhouette of a person is visible in the driver's seat of the car, seen through the side window.\\\midrule
\multicolumn{2}{l}{\textbf{Log-$\gamma$ 4-bits:} Similarity $6.2\pm2.99$ \quad Detail Level $5.8\pm1.50$ \quad Detail Match $4.5\pm3.00$ }&\multicolumn{2}{l}{\textbf{Log-$\gamma$ 4-bits:} Similarity      7.5 $\pm$ 0.71  \quad   Detail Level 7.0 $\pm$ 0.00  \quad   Detail Match 7.5 $\pm$ 0.71}\\\midrule
        \raisebox{-0.95\totalheight}{\includegraphics[width=0.3\linewidth]{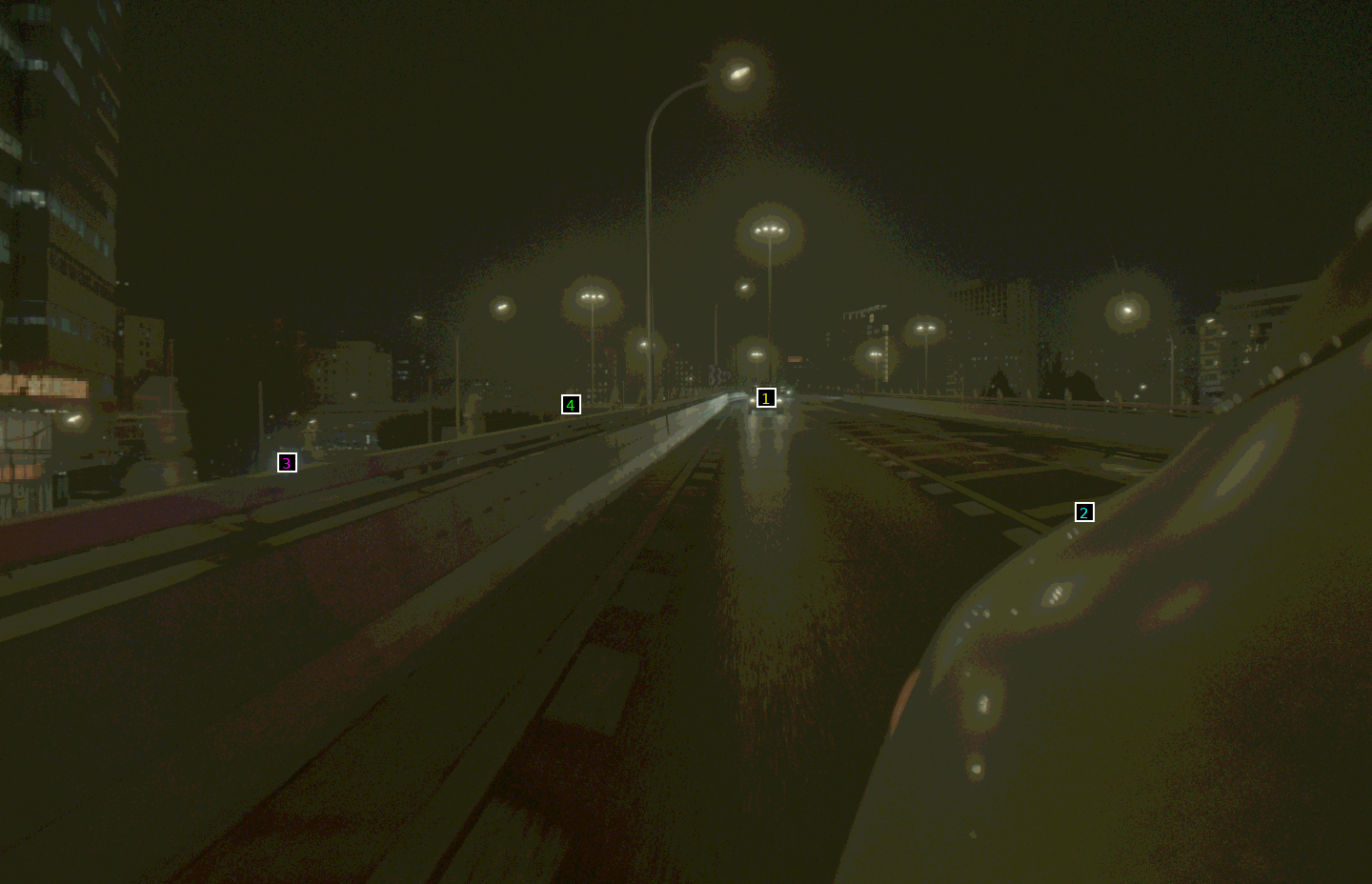}}&1. A light-colored car is in the distant center of the frame, driving on a wet road with its headlights on, which are reflecting on the dark pavement. \newline 2. A dark-colored car with a glossy finish occupies the bottom-right foreground of the image, with reflections from streetlights visible on its hood. \newline 3. A distant car with its lights on is partially visible on the far-left side of the road, behind a concrete barrier.\newline  4. A person is standing as a small, dark silhouette on the elevated walkway to the left of the road, behind a concrete barrier.
        &
        \raisebox{-0.95\totalheight}{\includegraphics[width=0.3\linewidth]{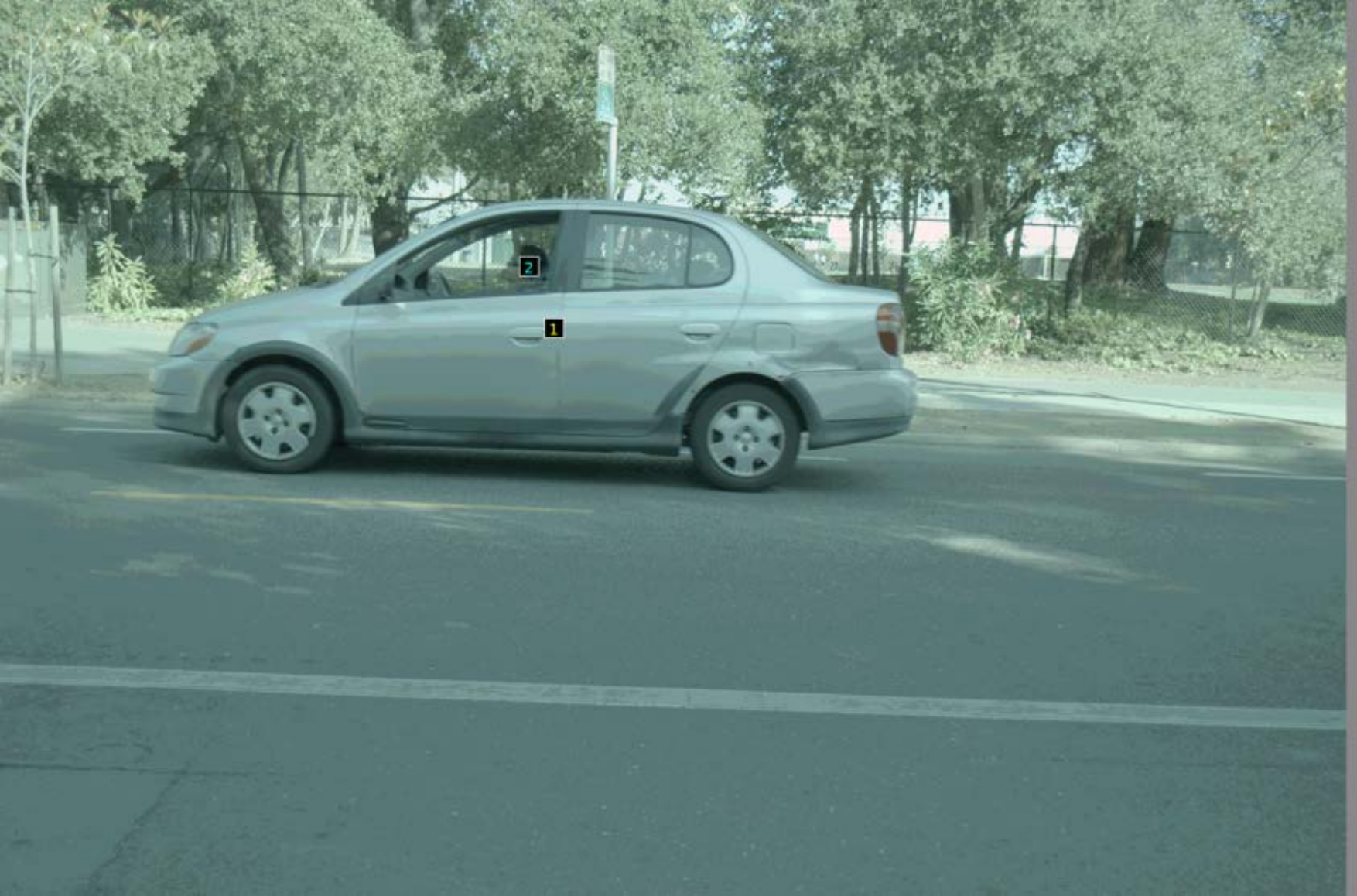}}&
1. A light silver four-door sedan is seen from the side, positioned in the middle of the frame and driving on an asphalt road. \newline 2. The head and shoulders of a person with dark hair are visible through the driver's side window of the silver car.
    \end{tabular}
    }
    \caption{Comparison of caption quality for different image representations, including quantitative scores. Captions generated from Log-$\gamma$ images provide more detail-level explanations compared to those from Linear 4-bit quantized images.}
    \label{fig:comparing_captions_1}
    \vspace{-1em}
\end{figure*}

Next, we study object description quality across different RAW preprocessing pipelines. 
Since current VLMs are closed and cannot be finetuned end-to-end on RAW, we adopt a zero-shot setting and reuse the $\gamma$ values that previously yielded strong robustness in our object detection experiments across architectures and bit depths. 
For this analysis, we randomly sample 500 images from the \rawdet{} validation set and treat high-resolution sRGB as our reference modality. 
Each image is annotated with set-of-marks and provided twice to Gemini-2.5-Pro~\cite{comanici2025_gemini-2.5} in order to probe the stability of its object descriptions on sRGB.

From the corresponding Bayer-patterned RAW measurements, we extract R, G, and B channels and construct two families of quantized RAW variants at 4, 6, and 8 bits: a linear mapping and a log + $\gamma$ mapping that combines logarithmic compression with the reused $\gamma$ values. In addition, we generate a 2$\times$ downsampled sRGB image whose effective resolution roughly matches that of the 8-bit RAW variants. 
All non-reference representations (downsampled sRGB, linear RAW, log + $\gamma$ RAW) are overlaid with the same set-of-marks and passed to Gemini-2.5-Pro for object descriptions. 
Caption quality is then evaluated against the high-resolution sRGB baseline using five metrics:Bleu, Regex Match, Similarity (1-10), Detail Match (1-10), and Detail Level (1-10), with the last three obtained from Gemini-2.5-Flash-Lite for consistent qualitative scoring. 
Formal metric definitions are deferred to the appendix.

The results in \cref{fig:reasoning} show a consistent pattern. 
Across all bit depths, linearly quantized RAW images produce captions that are less similar to the sRGB reference and clearly underperform in detail and detail match, indicating that naive RAW quantization discards semantically important structure. 
In contrast, log + $\gamma$ quantized RAW images retain much more fine-grained information and achieve scores close to, and sometimes approaching, high-resolution sRGB, while the 2$\times$ downsampled sRGB variant sits between the two extremes. 
This supports our central claim that careful RAW preprocessing can already narrow the gap between what an RGB-pretrained VLM can describe on sRGB and on RAW inputs. At the same time, the remaining gap and the strong sensitivity to preprocessing underline that bad RAW processing leads to systematically poorer descriptions, whereas good RAW-aware mappings are feasible and worth optimizing, as visualized in \cref{fig:comparing_captions_1}.
The object description track in \rawdet{} makes these trade-offs measurable and enables future work to benchmark RAW preprocessing pipelines, not only against sRGB, but also across multiple bit-depth-accurate RAW variants.

Looking forward, optimizing gamma specifically for image caption generation may further improve performance, potentially surpassing that of sRGB.

\section{Conclusion}
\rawdet{} fills a critical gap in the vision community by enabling robust, efficient, and generalizable object detection directly on RAW sensor data. 
By unifying and re-annotating four existing datasets, we provide a diverse and high-quality corpus that spans a wide range of lighting conditions, environments, and sensor types. Our dense, consistent annotations across seven common object categories resolve the limitations of prior RAW datasets, including noisy labels, limited class coverage, and inconsistent naming conventions. Moreover, the addition of object captions makes it a useful tool for bridging low-level RAW image processing with high-level semantic understanding, enabling tasks like caption-guided detection on RAW data.

\rawdet{} supports controlled benchmarking under various quantized inputs, reflecting practical constraints at the sensor level, such as reduced bandwidth and energy consumption. We empirically demonstrate that training on \rawdet{} leads to consistent performance improvements across standard detectors like Faster R-CNN, RetinaNet, and PAA. Moreover, we show that even a LVLM, MM-Grounding DINO, can make meaningful zero-shot predictions on quantized RAW inputs, particularly when simple input normalization strategies are applied. 
We anticipate that \rawdet{} will serve as a catalyst for further research in low-level vision, quantization-aware modeling, and end-to-end co-design of sensors and modern vision systems.

\vspace{0.1cm}
\noindent\textbf{Limitations. } 
\rawdet{} improves coverage of sensor settings by building on diverse prior datasets, but its coverage of the broader RAW sensor landscape remains limited. The dataset is also imbalanced across sensors and bit depths, with most images at 24-bit. Even if the practical impact is unclear, a more balanced sensor representation is desirable. In principle, additional RAW datasets could be merged into \rawdet{} and annotated using our pipeline, but expanding annotations is costly because high-quality annotation tools (including ours) are paid.

\noindent\textbf{Acknowledgements. } The authors acknowledge support by the DFG Research Unit 5336 - Learning to Sense (L2S). We would also like to thank Penelope Natusch for her helps with the project. The authors gratefully acknowledge the computing time provided on the high-performance computer HoreKa by the National High-Performance Computing Center at KIT (NHR@KIT).
We thank Janis Keuper for his support.


\newpage

\clearpage
{
    \small
    \bibliographystyle{ieeenat_fullname}
    \bibliography{main}
}
\newpage


   

\onecolumn
\maketitleocsupplementary
\setcounter{page}{1}

\appendix

\renewcommand{\contentsname}{Appendix Contents}
\startcontents[appendix]
\printcontents[appendix]{l}{1}{\setcounter{tocdepth}{2}}

\section{Experimental Details}
\label{appendix:experimental_details}
Following, we provide more details regarding the object detection experimental setup: training details for the architecture and details on using the foundation model for generating the new annotations for \rawdet{}.
\subsection{Object Detection Training Details}
Following previous works, for training traditional object detection models, we follow multi-scale training where the shorter image side is scaled to one of the sides randomly selected from a set of sides: [480, 512, 544, 576, 608, 640, 672, 704, 736, 768, 800] using nearest neighbor interpolation and the longer side is scaled to maintain the aspect ratio. 
We apply a warm-up for the first 1000 iterations, linearly increasing the learning rate (lr) from  $1e^{-3}$ to $2.5e^{-3}$ followed by a multi-step learning rate scheduler. 

\subsection{Details of Foundation Model used for Annotations}
\label{appendix:annotating_model_details}
We used the paid API version at: \url{https://deepdataspace.com/request_api}.
We use their model ``DetectionModel.GDino1\_5\_Pro'' for generating the bounding boxes and classes such that we only annotate objects with 20\% or less overlap with another object of the same class (following recommended settings).
The text prompt we used for generating the annotations was ``car \onedot truck \onedot tram \onedot person \onedot bicycle \onedot motorcycle \onedot bus \onedot''.
Each query to the API costs $\yen$0.1, thus, including the debugging and ablations over prompts and classes, the total cost was $\yen$3402.80.
Finally, after obtaining the annotations, we observed that the model was often hallucinating low-confidence objects.
Thus, we filter out these hallucinated objects, after carefully observing multiple annotations and confidences, by using a confidence score threshold of 0.8, \ie only annotations with confidence scores of 0.8 and above were retained, and the remaining were filtered out.

\section{User Study Comparing Our New Object Detection Annotations to Old}
\label{appendix:user_study}
We did a user survey for our new data annotations and received 53 responses. 
For the survey, we collected 100 random pairs of annotations such that for each image chosen at random, we had the annotation from \rawdet{} and the annotation provided by the original dataset. 
We split the 100 random pairs of annotations into 4 sets of 25 questions each.
Each question shuffled the order of option A and option B, with annotation from \rawdet{} for that image being either option A or option B, and the other option being the annotation from the original dataset.



Given 25 questions to each user, and since 53 users responded, we had exactly 1325 instances of comparisons between the \rawdet{} annotations and the original annotations answered.
From these 1325 instances, users preferred the \rawdet{} annotations in 1002 instances.
That is, \textbf{users preferred \rawdet{} annotations for 75.62\% of all the instances}.
This clearly demonstrates the improvement of the annotations provided by \rawdet{}.
For ease of access to the comparisons, in \Cref{appendix:vis}, we extended the comparison from \Cref{fig:teaser} and show a few examples comparing the annotations from \rawdet{} and the original annotations. 

\begin{figure}
    \centering
    \includegraphics[width=1.0\linewidth]{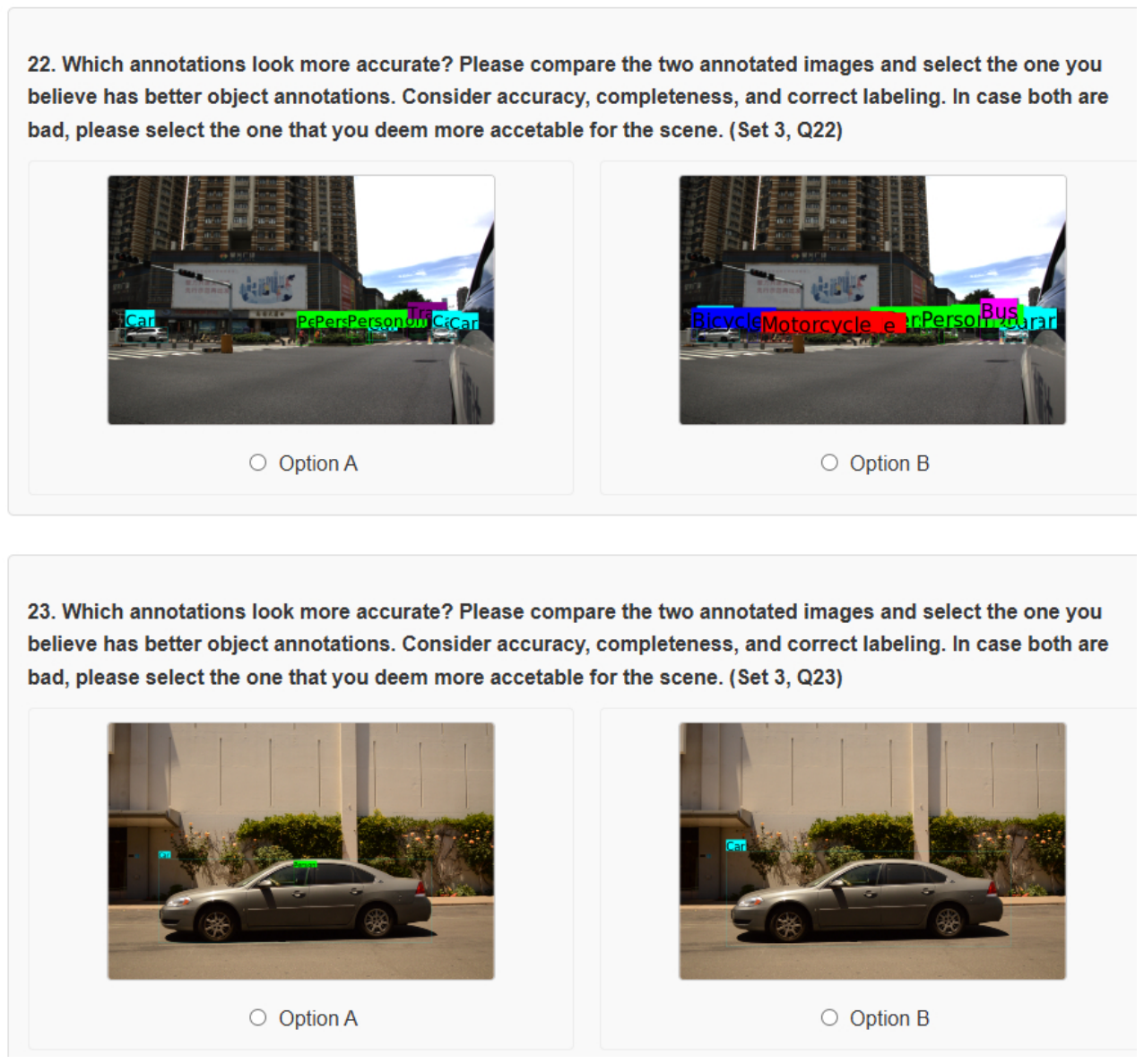}
    \caption{Examples from the annotation-quality user study, where participants inspected bounding boxes and compared newly generated annotations against older versions to assess improvements in quality.}
    \label{fig:user_study_anno}
\end{figure}

\paragraph{Limitations Of The User Study. }
Post the study, a few users informed that they had to zoom in quite a bit to confirm that one annotation is better than the other.
We suspect that the annotations that required zooming in were the annotations from \rawdet{} since it provides very fine-grained detections as well.
Users reported that, in a few instances, those who are not vigilant may not choose the \rawdet{} annotations, and thus the final scores might underrepresent the benefit of the annotations from \rawdet{}.

\section{Visualizing And Comparing Annotations From \rawdet}
\label{appendix:vis}
\begin{figure}
    \centering
    \resizebox{\linewidth}{!}{
   \begin{tabular}{@{}lc@{\hspace{1.5mm}}c@{\hspace{1.5mm}}c@{\hspace{1.5mm}}c@{\hspace{1.5mm}}c@{}}
         & \textbf{PASCAL RAW} & \textbf{NOD-Nikon} & \textbf{NOD-Sony} & \textbf{RAOD-Day} & \textbf{RAOD-Night} \\
         \rotatebox{90}{\phantom{space}\textbf{Original}} 
         &
         \includegraphics[width=3.4cm, height=2.8cm]{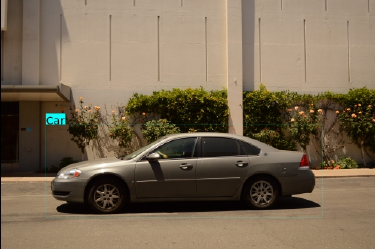}
         &
         \includegraphics[width=3.4cm, height=2.8cm]{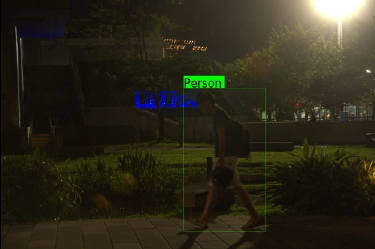}
         &
         \includegraphics[width=3.4cm, height=2.8cm]{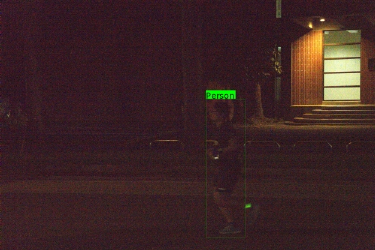}
         &
         \includegraphics[width=3.4cm, height=2.8cm]{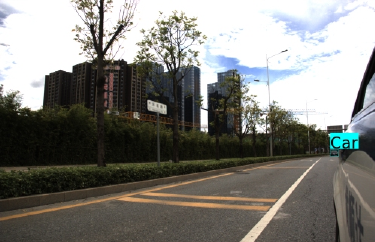}
         &
         \includegraphics[width=3.4cm, height=2.8cm]{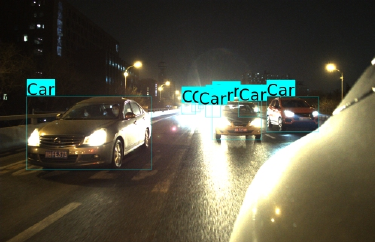}
         \\

         \rotatebox{90}{\phantom{}\textbf{\rawdet{} (Ours)}}
         &
         \includegraphics[width=3.4cm, height=2.8cm]{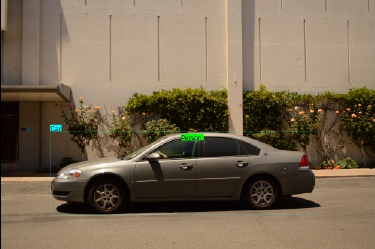}
         &
         \includegraphics[width=3.4cm, height=2.8cm]{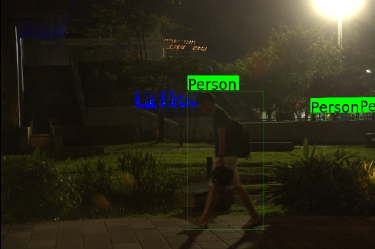}
         &
         \includegraphics[width=3.4cm, height=2.8cm]{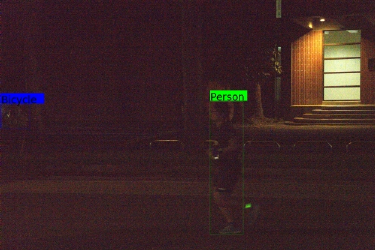}
         &
         \includegraphics[width=3.4cm, height=2.8cm]{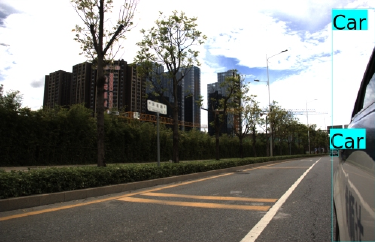}
         &
         \includegraphics[width=3.4cm, height=2.8cm]{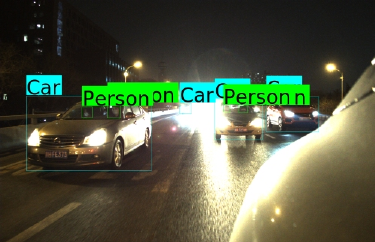} \\


\rotatebox{90}{\phantom{space}\textbf{Original}} 
         &
         \includegraphics[width=3.4cm, height=2.8cm]{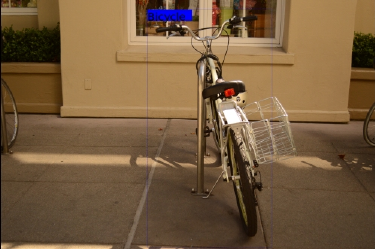}
         &
         \includegraphics[width=3.4cm, height=2.8cm]{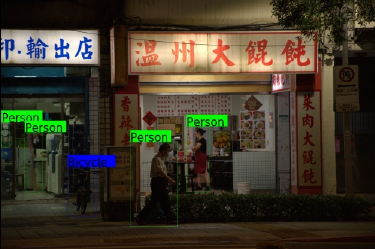}
         &
         \includegraphics[width=3.4cm, height=2.8cm]{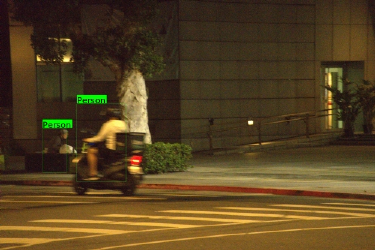}
         &
         \includegraphics[width=3.4cm, height=2.8cm]{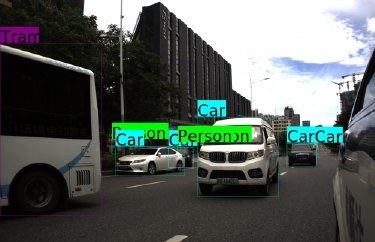}
         &
         \includegraphics[width=3.4cm, height=2.8cm]{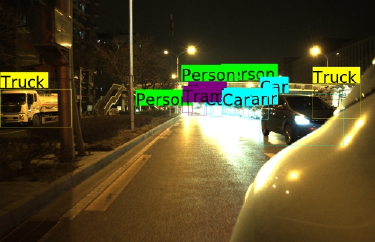}
         \\

         \rotatebox{90}{\phantom{}\textbf{\rawdet{} (Ours)}}
         &
         \includegraphics[width=3.4cm, height=2.8cm]{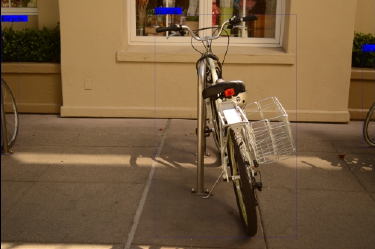}
         &
         \includegraphics[width=3.4cm, height=2.8cm]{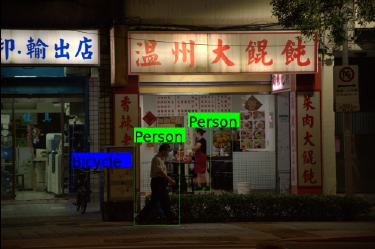}
         &
         \includegraphics[width=3.4cm, height=2.8cm]{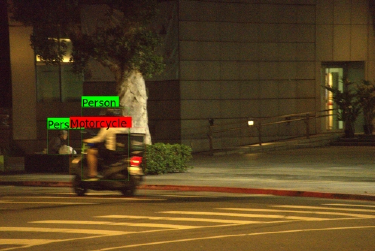}
         &
         \includegraphics[width=3.4cm, height=2.8cm]{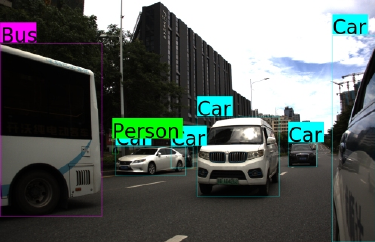}
         &
         \includegraphics[width=3.4cm, height=2.8cm]{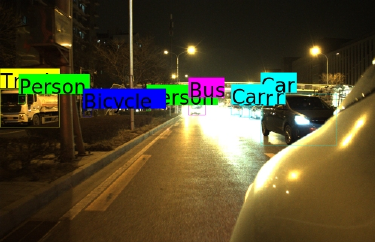} \\


\rotatebox{90}{\phantom{space}\textbf{Original}} 
         &
         \includegraphics[width=3.4cm, height=2.8cm]{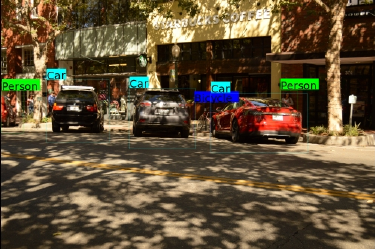}
         &
         \includegraphics[width=3.4cm, height=2.8cm]{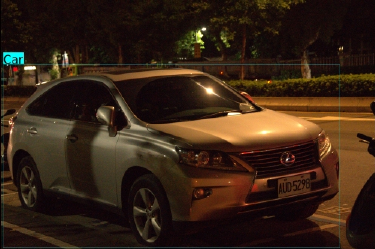}
         &
         \includegraphics[width=3.4cm, height=2.8cm]{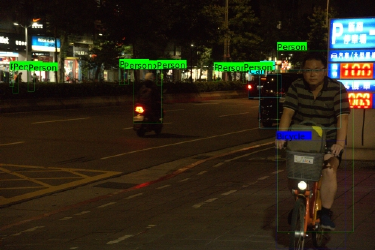}
         &
         \includegraphics[width=3.4cm, height=2.8cm]{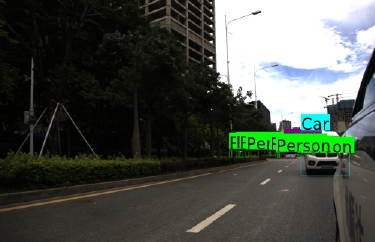}
         &
         \includegraphics[width=3.4cm, height=2.8cm]{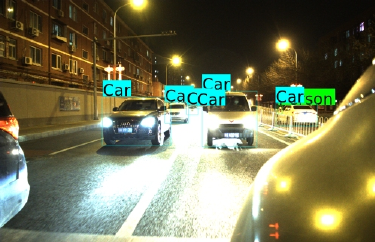}
         \\

         \rotatebox{90}{\phantom{}\textbf{\rawdet{} (Ours)}}
         &
         \includegraphics[width=3.4cm, height=2.8cm]{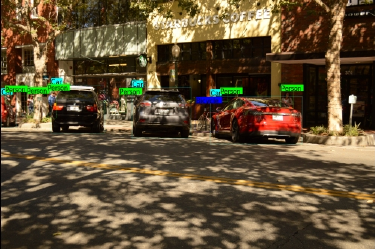}
         &
         \includegraphics[width=3.4cm, height=2.8cm]{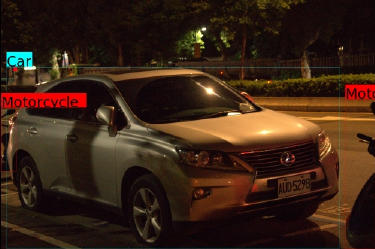}
         &
         \includegraphics[width=3.4cm, height=2.8cm]{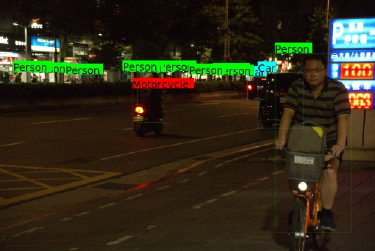}
         &
         \includegraphics[width=3.4cm, height=2.8cm]{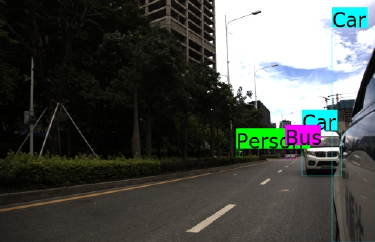}
         &
         \includegraphics[width=3.4cm, height=2.8cm]{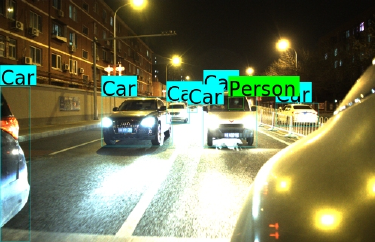} \\


\rotatebox{90}{\phantom{space}\textbf{Original}} 
         &
         \includegraphics[width=3.4cm, height=2.8cm]{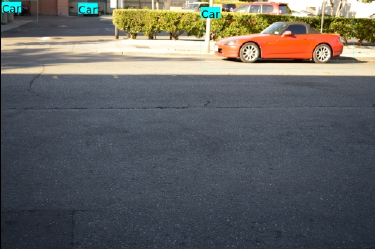}
         &
         \includegraphics[width=3.4cm, height=2.8cm]{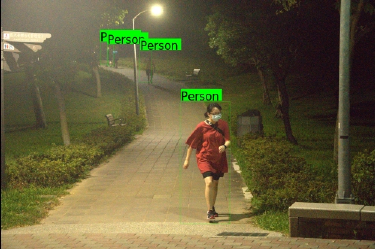}
         &
         \includegraphics[width=3.4cm, height=2.8cm]{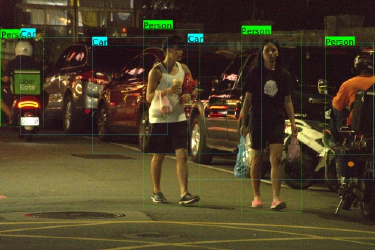}
         &
         \includegraphics[width=3.4cm, height=2.8cm]{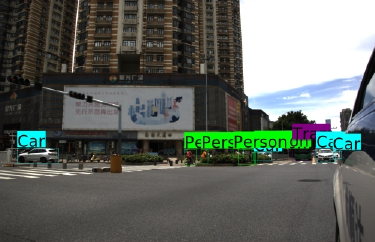}
         &
         \includegraphics[width=3.4cm, height=2.8cm]{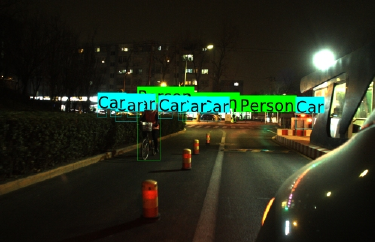}
         \\

         \rotatebox{90}{\phantom{}\textbf{\rawdet{} (Ours)}}
         &
         \includegraphics[width=3.4cm, height=2.8cm]{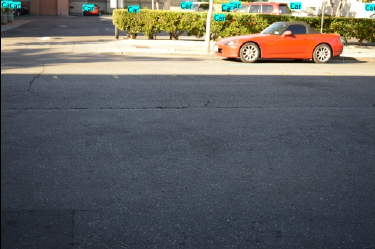}
         &
         \includegraphics[width=3.4cm, height=2.8cm]{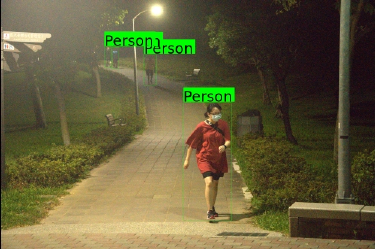}
         &
         \includegraphics[width=3.4cm, height=2.8cm]{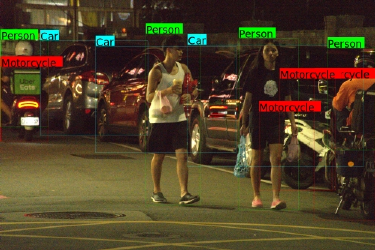}
         &
         \includegraphics[width=3.4cm, height=2.8cm]{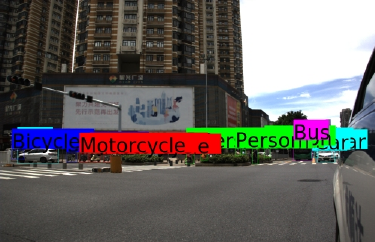}
         &
         \includegraphics[width=3.4cm, height=2.8cm]{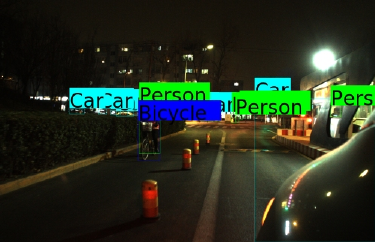} \\
         
    \end{tabular}
    }
    \caption{Comparing ground truth annotations provided in the original datasets and the new ones proposed in \rawdet{} as used in the user study. Please note that the visualizations here are heavily compressed due to size limitations. 
    }
    \label{fig:appendix:comparing_annotations}
\end{figure}

In \Cref{fig:appendix:comparing_annotations}, we observe that \rawdet{} covers many small objects in the images that were originally missing from the annotations, and correctly labels many misclassified objects in the original annotations.

\section{Additional Dataset Statistics}
\label{appendix:data}

\begin{table*}[t]
\centering
\caption{Characteristics of Annotated Objects in \rawdet{}.}
\begin{tabular}{lccccccc}
\toprule
  &  Bicycle & Bus & Car & Motorcycle & Person & Tram & Truck \\ 
 \midrule
Avg. Obj Size (H $\times$ W) & 477 $\times$ 431 & 224 $\times$ 301 & 250 $\times$ 257 & 231 $\times$ 195 & 508 $\times$ 213 & 177 $\times$ 307 & 231 $\times$ 257 \\
Avg. Obj Count/Image & 1.95 & 1.28 & 7.63 & 2.12 & 3.62 &1.12 & 1.53\\
Avg. Aspect Ratio & 0.939 & 1.486 & 1.350 & 0.888 & 0.528 &2.192&1.197 \\
\bottomrule
\end{tabular}
\label{tab:object_statistic}
\end{table*}

In \autoref{tab:object_statistic}, we list different object characteristics in the proposed \rawdet dataset. Average object size is computed by averaging the height and width of bounding boxes across all images in the train and validation sets for a particular category. We observe that the bicycle category has the largest object size, and the motorcycle, truck, and tram have the smallest. The category "car" has the highest average object count per image, which is intuitive due to the outdoor nature of the proposed dataset. Average aspect ratio is computed by calculating the aspect ratio (width/height) of every object in a category and averaging over the entire dataset.  

\section{Additional Benchmarking Details \eg details on the ISP used for RAW RGB conversion of the images from the RAOD dataset.}
\label{appendix:benchmark}
Except for the Zurich Dataset, the remaining individual datasets have a Bayer pattern of RGGB, whereas Zurich raw images have an RGBG pattern. We extract red, green, and blue channels from the relevant Bayer pattern and take the mean of the double green channels. For quantization experiments, we use \textit{torch.floor} as a quantization function. In the case of gamma scaling, we use a straight-through estimator to let the gradients pass through the quantization operation. The RGB images of all the individual datasets are publicly provided along with the raw images, except for the RAOD dataset.
So, we generated them by first extracting the Red, Green,
and Blue channels from the original raw image, followed by a gray
world white balance algorithm. Then, we apply gamma
correction, where the value of gamma is chosen by hit-and-run to be 0.09. 

\section{Advantage of Consolidated Dataset}
\label{appendix:consolidated}
\begin{figure}[t]
    \centering
    \includegraphics[width=1.0\linewidth]{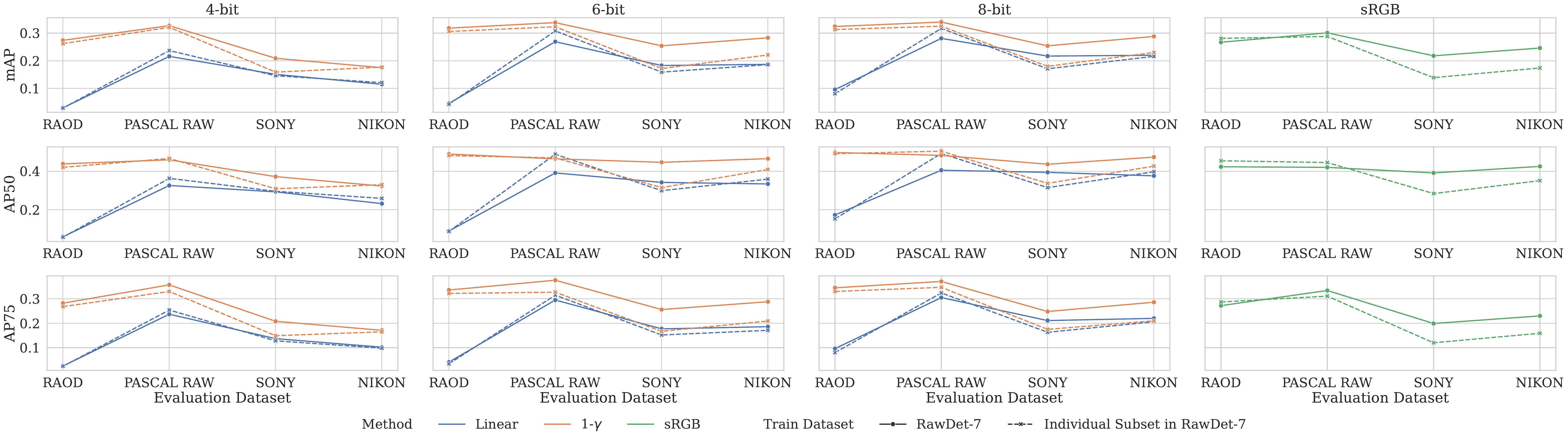}
    \caption{Results with PAA for training on combined \rawdet{} when evaluated on each subset, vs.~training only on a subset and evaluated on the subset's test set. Combining improves results on the majority of the subsets in the case of gamma scaling and sRGB. These comparisons include all metrics (mAP, AP50, AP75) and cover 4, 6, and 8-bit depth data as well as sRGB.}
    \label{fig:combined_individual_training_paa}
\end{figure}

We show the advantage of adding all the individual datasets together for training vs. training only with the individual datasets in \autoref{fig:combined_individual_training_paa}  for PAA architecture. We observe that while training with linear quantization without any scaling results in no performance gain when trained in a combined fashion, gamma scaling and RGB experiments usually benefit from combined training. This is especially visible in the case of the Sony and Nikon Datasets. 

\begin{figure}[t]
    \centering
    \includegraphics[width=1.0\linewidth]{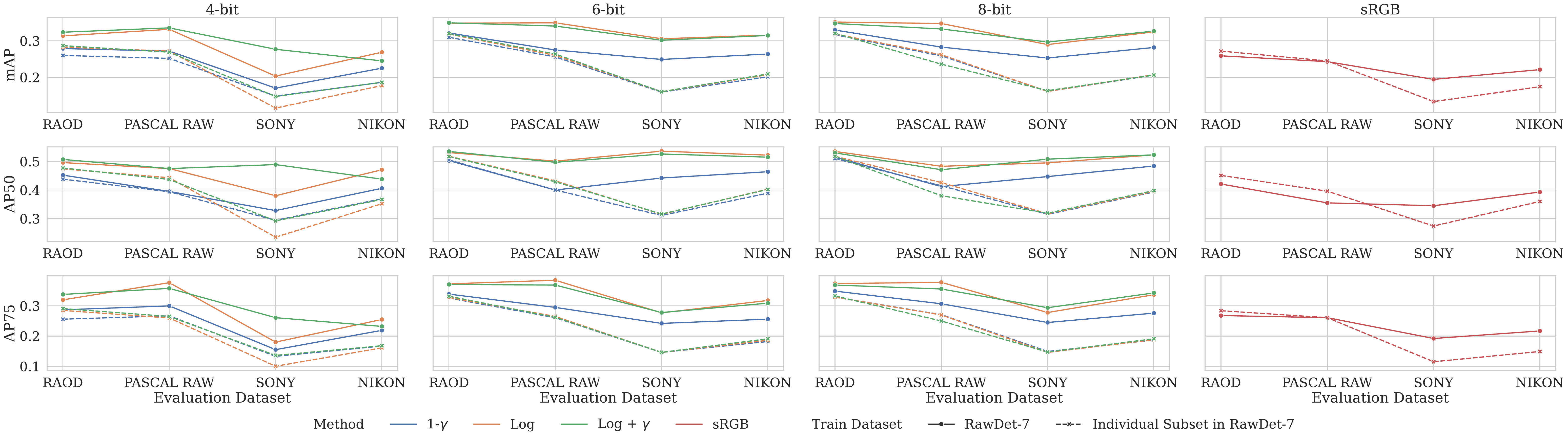}
    \caption{Results with FRCNN for training on combined \rawdet{} when evaluated on each subset, vs.~training only on a subset and evaluated on the subset's test set. Combining improves results on the majority of the subsets in the case of gamma scaling and sRGB. These comparisons include all metrics (mAP, AP50, AP75) and cover 4, 6, and 8-bit depth data as well as sRGB. }
    \label{fig:combined_individual_training_frcnn}
\end{figure}

In \autoref{fig:combined_individual_training_frcnn}, we show results with FRCNN for training on combined \rawdet{} when evaluated on each subset, vs.~training only on a subset and evaluated on the subset's test set. Combining improves results on the majority of the subsets in the case of gamma scaling and sRGB. These comparisons include all metrics (mAP, AP50, AP75) and cover 4, 6, and 8-bit depth data as well as sRGB.

\begin{figure}
    \centering
    \includegraphics[width=1.0\linewidth,trim={2cm 0cm 2cm 0cm},clip]{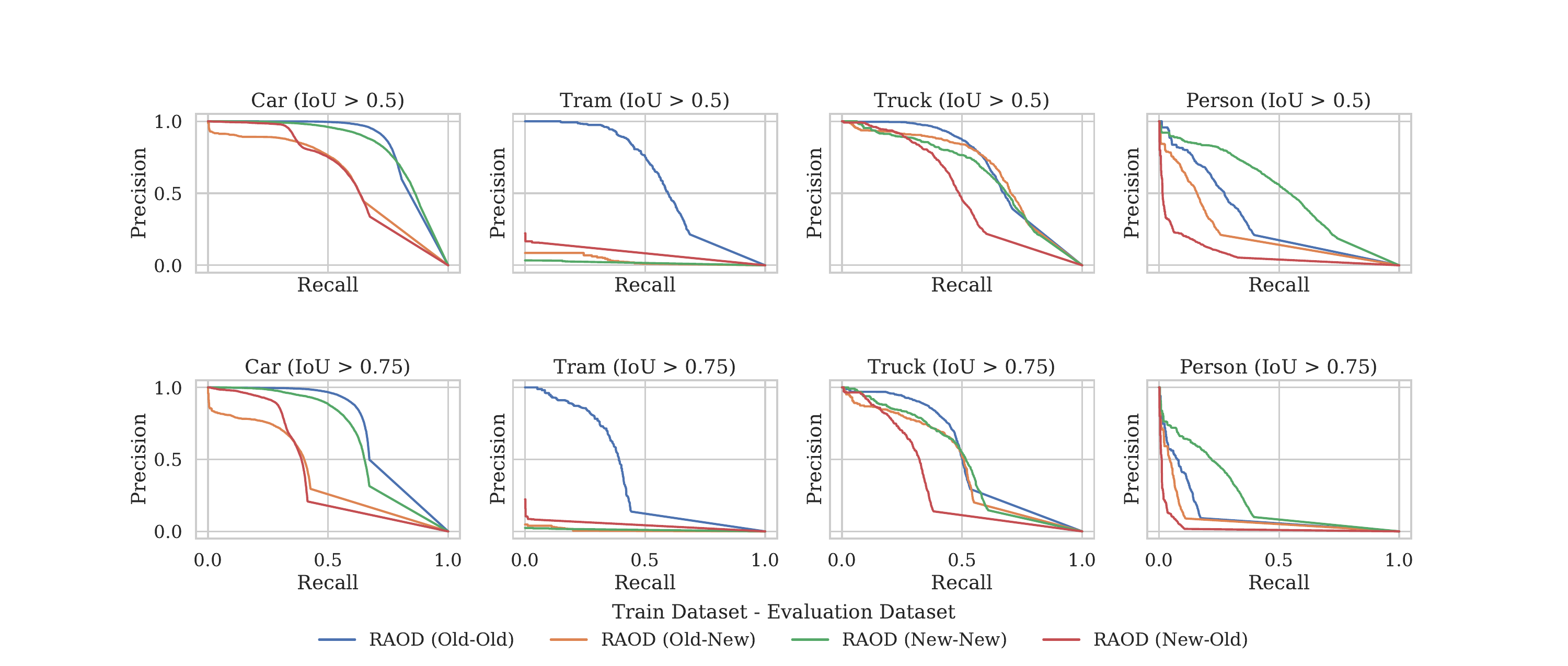}
    \caption{Precision Recall Curves for RAOD. ``Old'' refers to the annotations as proposed in the original dataset, whereas ``New'' means the annotations in \rawdet{}. All the models are trained on FasterRCNN for 8-bit quantization and jointly learnt with 1 gamma.
    As mentioned in the legend, the keys are ``(Train Dataset - Evaluation Dataset)''. Curves are shown for the classes car, truck, tram, and person.}
    \label{fig:pr_raod_supp}
\end{figure}

In \autoref{fig:pr_raod_supp}, we show precision-recall curves for the RAOD dataset. ``Old'' refers to the annotations as proposed in the original dataset, whereas ``New'' refers to the annotations in \rawdet{}. All the models are FasterRCNNs, trained for 8-bit quantization and jointly learnt with 1 gamma. As mentioned in the legend, the keys are ``(Train Dataset - Evaluation Dataset)''. Curves are shown for the classes car, truck, tram, and person. 

\begin{figure*}
    \centering
    \includegraphics[width=1.0\linewidth]{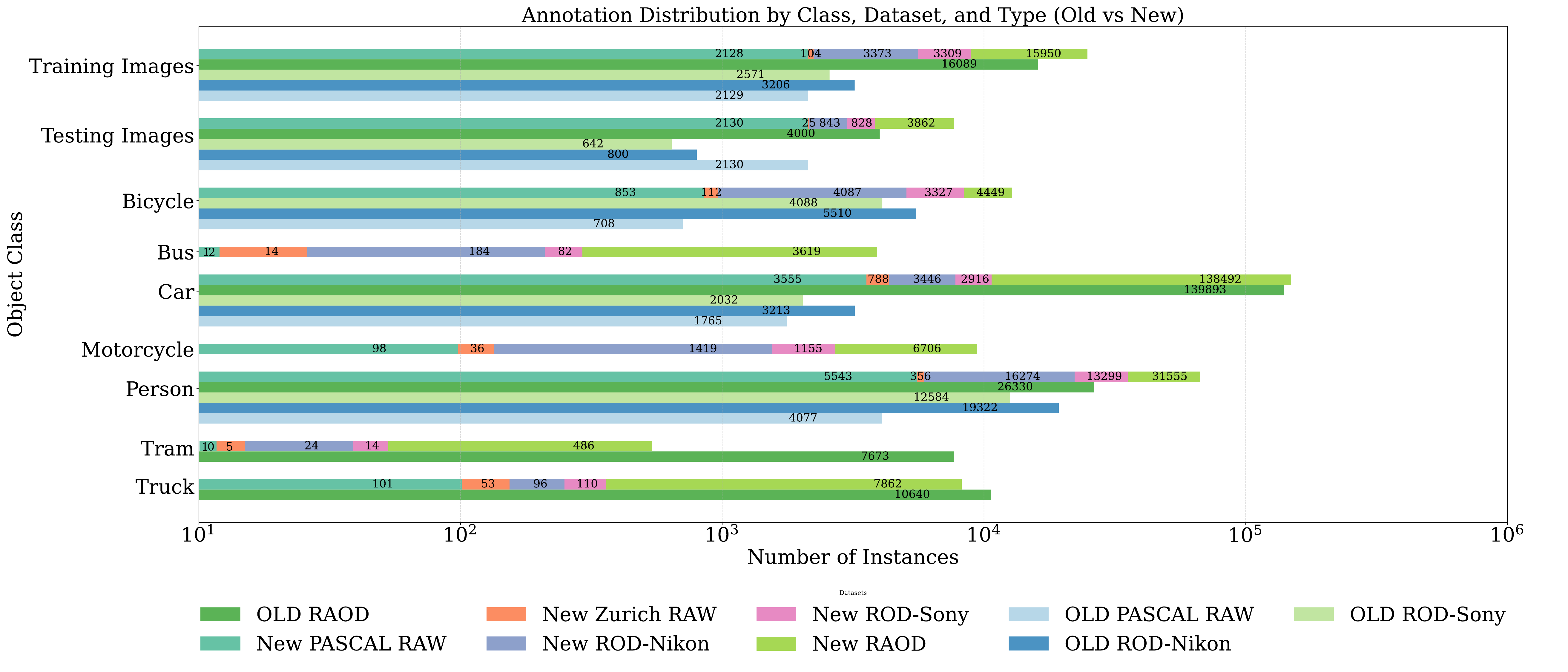}
    \caption{Dataset Statistics: Breakdown of instances in the original proposed datasets indicated by `OLD' and the re-annotated images for those respective datasets indicated by `NEW' for the classes in \rawdet{}. The proposed \rawdet{} is a consolidation of RAW input images from the `OLD' datasets provided with the `NEW' annotations. For some classes, it appears that `OLD RAOD' has more instances than the proposed \rawdet{}. However, as shown in \cref{fig:teaser}, the original annotations of the RAOD dataset contain hallucinations for classes like `Tram', `Truck', and `Car'.}
    \label{fig:dataset_statistics}
\end{figure*}
In \autoref{tab:sensor_statistic}, we summarize the statistics on the different datasets that we consolidate into \rawdet{}. The individual datasets have each a particular sensor model and bit depth. The combined dataset, therefore, provides significantly increased data diversity.


\begin{table*}[t]
\centering
\caption{Characteristics of Different Datasets in \rawdet{}.}
\resizebox{\linewidth}{!}{
\begin{tabular}{lccccc}
\toprule
 & Pascal Raw & Zurich & Raw-NOD-Nikon & Raw-NOD-Sony & RAOD \\ 
 \midrule
Sensor Model & Nikon D3200 DSLR & MP Sony Exmor IMX380 & Nikon D750 & Sony RX100 VII & Sony IMX490 \\
Resolution & 6034 × 4012& 2944 x 3958 & 3968 × 2640 & 4256 × 2848 & 2880 × 1856 \\
Bit Depth & 12 & 10 & 14 & 14 & 24 \\
Scnearios & Day& Day & Night & Night & Day and Night \\
\bottomrule
\end{tabular}
}
\label{tab:sensor_statistic}
\end{table*}

In \autoref{fig:dataset_statistics}, we provide detailed dataset statistics. In particular, we report statistics on instances in the original proposed datasets indicated by ‘OLD’ and the re-annotated images for
those respective datasets indicated by ‘NEW’ for the classes in \rawdet{}. The proposed \rawdet{} is a consolidation of RAW input
images from the ‘OLD’ datasets provided with the ‘NEW’ annotations. For some classes, it appears that ‘OLD RAOD’ has more instances
than the proposed \rawdet{}. Note that ROAD has a large number of objects annotated. However, as shown in \autoref{fig:teaser}, the original annotations of the RAOD dataset contain hallucinations for
classes like ‘Tram’, ‘Truck’, and ‘Car’.

The consolidated dataset \rawdet{} provides cleaned labels for the diverse RAW data samples.

\section{Object Descriptions}
\label{appendix:object_description}

\subsection{Generating Object Descriptions}
For object captioning, we use the \texttt{caption prompt used for object-level descriptions} prompt with \texttt{Gemini-2.5-Pro}. 
The model receives the overlaid image where each detected object is marked with a numbered square, together with the list of corresponding class labels. 
The prompt enforces a strict line-based output format of the form ``\texttt{<number>: <caption>}'' and requires exactly one caption per numbered object, in order from 1 to $\mathrm{N}$. We call \texttt{Gemini-2.5-Pro} on each marked image (twice for the sRGB reference and once for each RAW or RGB-downsampled variant) and then parse the returned text using a regular expression to obtain a dictionary that maps object indices to their captions, which is then used for both analysis and visualization.
\newpage
\subsubsection{Prompts for Generating Object Descriptions}
\begin{promptbox}[Caption prompt used for object-level descriptions]
You will be shown a photo with multiple objects with these 7 classes of interest: cars, truck, tram, person, bicycle, motorcycle, bus.
Each object of interest is marked by a black square containing a coloured number.

Task: For EACH object with a marked number, produce ONE detailed, factual caption that ONLY describes that numbered object.
Include concrete details when visible: color, material/texture, approximate size class, local position (e.g., top-left),
distinctive parts, visible text/markings, and immediate context relations.

STRICT OUTPUT:
Write ONE line per object using this exact pattern: <number>: <caption>
Acceptable separators after the number are ':', ')', '.', or '-' (e.g., '1: ...' or '1) ...').
Do NOT include headings, bullets, blank lines, or any extra text before/after the list.
If a numbered object cannot be seen or identified, write exactly: 'I cannot see the object'.
The numbering starts at 1 and goes up to N, where N is the total number of marked objects in the image.
There should be exactly N lines in your output, one for each numbered object.
If N is the highest number visible, ensure you include captions for all numbers from 1 to N in order.
We will tell you the object class of each marked object in the following way:
1: <CLASS>
2: <CLASS>
...
N: <CLASS>

If a numbered mark itself cannot be seen or identified in the image, write exactly: 'I cannot see the mark'.
DO NOT HALLUCINATE ANY MARKED NUMBERS.
DO NOT TALK ABOUT THE MARK SQUARES OR NUMBERS ITSELF or the WHOLE IMAGE.
\end{promptbox}

\begin{figure}
    \centering
    \includegraphics[width=1.0\linewidth, trim=0cm 4cm 0cm 4cm, clip]{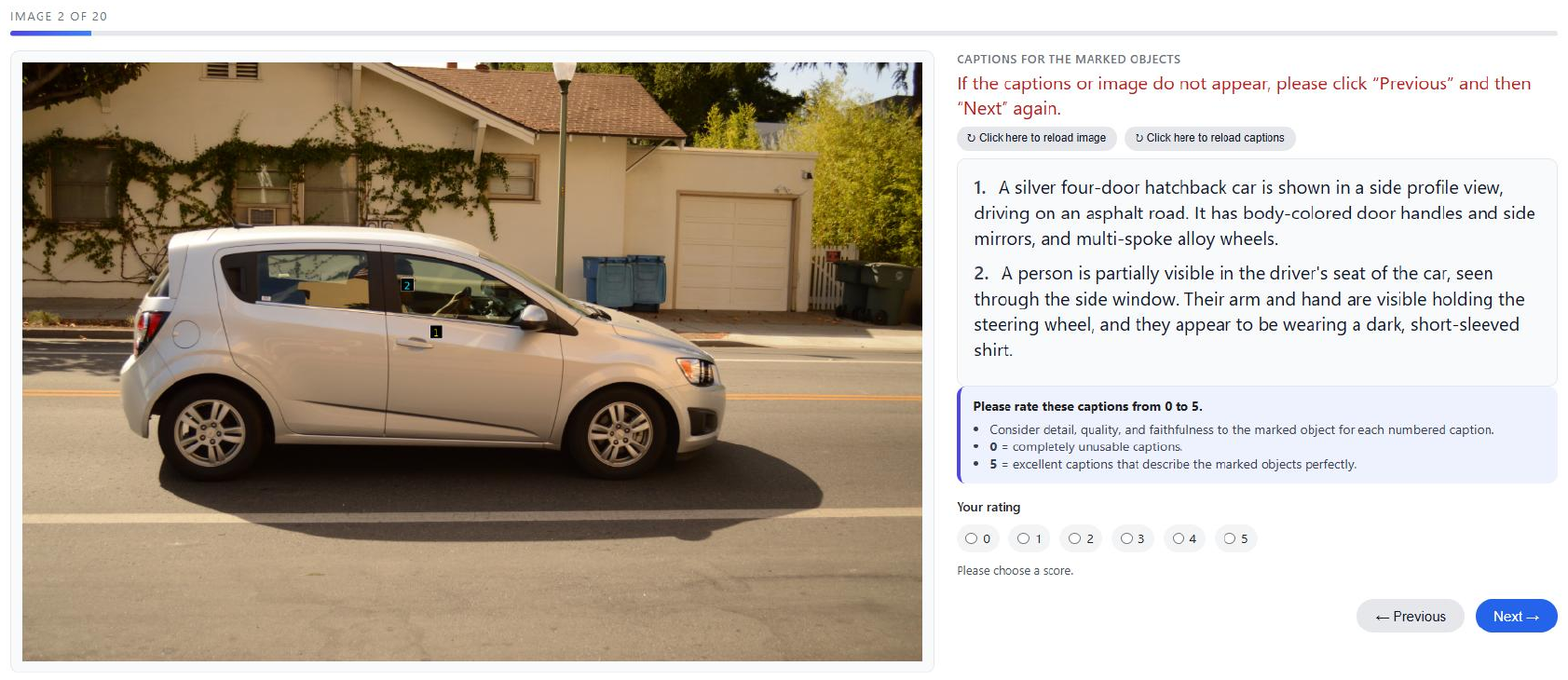}
    \caption{Example from the user study used to assess caption quality. Participants were shown images with designated markers and asked to rate the quality of the accompanying captions, where 0 is completely unusable and 5 is fully descriptive, clear, and highly informative.}
    \label{fig:user_study_caption}
\end{figure}

\subsection{User Study to Validate Quality of the Object Descriptions}
\label{appendix:user_study_description} We conducted a user study to evaluate the effectiveness of generated captions. We selected 100 images with the highest mean detail levels and then randomly sampled 20 images and their respective captions from these for each user study. 
We chose the 100 images with the highest mean detail level, since usually it is the longer output text where generative models hallucinate the most, as they often try to make up details.
We wanted the user study to validate that this is not the case in our proposed object descriptions.
Participants were shown each image along with its captions and asked to rate the quality of the captions on how well they described the marked objects in each image on a scale from 0 to 5.
Here, 0 meant that the object description does not describe the object at all, and 5 meant that the object description describes the respective marked object reasonably well, leaving out almost no describable visual cues.

A total of 63 users participated in this study, so a total of 63$\times$20, that is, 1260 samples were collected for the quality of the captions.
We received an average rating of $\frac{4.07\pm0.48}{5}$, that is $\approx$82\% score, which demonstrates the good quality of the object descriptions generated.

\subsection{Metrics for Evaluating Object Descriptions}
\label{appendix:metrics}

\begin{itemize}
    \item \textbf{BLEU}: Measures how much the generated text overlaps in wording with a reference.
    \item \textbf{Regex Match}: Checks if the output matches a specific pattern or format.
    \item \textbf{Similarity (1–10)}: Rates how semantically similar the meaning of two texts are, irrespective of the specific details. 
    \item \textbf{Detail Level (0–10)}: Rates how much detail or specificity the output contains.
    \item \textbf{Detail Match (1–10)}: Rates how well specific details in the output match the reference.
    
\end{itemize}

\paragraph{Similarity. }
To quantify semantic similarity between two captions that describe the same object, we use the \texttt{Similarity scoring prompt} with \texttt{Gemini-2.5-Flash-Lite}. 
This prompt presents Caption A and Caption B and instructs the model to ignore surface-level differences such as wording (for example, ``car'' versus ``sedan'') and colour or brightness variations, and to focus instead on whether both captions refer to the same underlying object with consistent meaning. 
The model is required to output a single integer in [1, 10] on a separate line, where 1 denotes unrelated or contradictory descriptions and 10 denotes near-identical semantics. 
We use this score as the \textit{Similarity} metric for each caption pair, where one of the captions is always the first caption generated for the full-scale RGB image.

\paragraph{Detail Level. }
To assess how informative a single caption is, we use the \texttt{Detail level prompt} with \texttt{Gemini-2.5-Flash-Lite}. The prompt asks the model to rate the richness of grounded detail in one caption only, again returning a single integer score in [0, 10]. The instructions explicitly direct the model to base its judgment on concrete attributes such as materials, positions, parts, readable text, and local relations, while ignoring fluency and style. 
A score of 0 indicates no grounded detail, whereas 10 corresponds to a very specific, highly descriptive caption. 
We treat this as the \textit{Detail Level} metric and compute it for each caption that we evaluate.

\paragraph{Detail Match. }
Finally, we measure the consistency of fine-grained information between two captions using the \texttt{Detail match prompt}, again with \texttt{Gemini-2.5-Flash-Lite}. 
Here, Caption A (first full-scale sRGB caption) acts as the reference and Caption B as the candidate, and the model is instructed to assign a score in [1, 10] based on the overlap of concrete, verifiable details such as materials, positions, parts, readable text and relations, while explicitly disregarding differences in colours or brightness due to varying image processing. 
Low scores correspond to almost no shared details, whereas scores near 10 indicate high overlap in grounded content. This yields the \textit{Detail Match} metric, which we aggregate over objects and variants to analyze how well different pipelines preserve detailed object-level information.
\vfill

\subsubsection{Prompts for LLM as a Judge}
\begin{promptbox}[Similarity scoring prompt]
You are scoring the semantic similarity between two captions that describe the SAME object.
Check if they are roughly talking about the same object, ignore details.
Disregard similarity in colour, and disregard difference in brightness of the colours; disregard the exact word used, for example, car or sedan; focus on overall meaning match.
Return ONLY one integer in [1,10] on a single line with no other text.
Scoring: 1=unrelated/contradictory, 10=near-identical meaning (paraphrases penalize contradictions).
Be strict about the overall meaning match.

Caption A:
{A}

Caption B:
{B}
\end{promptbox}

\begin{promptbox}[Detail level prompt]
You are scoring the richness of a SINGLE caption, that is, how long and detailed and descriptive it is.
Return ONLY one integer in [0,10] on a single line with no other text.
Score based on concrete grounded attributes (colors, materials, positions, parts, readable text, relations).
Be strict about verifiable details only. 10=very rich and specific, 0=no grounded detail.
Do not look for the length of the caption, only the amount of concrete details.
Ignore fluency/style. 0=no grounded detail, 10=very rich and specific.

Caption:
{C}
\end{promptbox}

\begin{promptbox}[Detail match prompt]
You are scoring the overlap of concrete details between two captions.
Return ONLY one integer in [1,10] on a single line with no other text.
Score high only if both share consistent verifiable details (materials, positions, parts, readable text, relations).
Disregard colour differences, and difference in brightness of the colours as the images are processed differently, and colour differences are expected, so disregard them.
Ignore fluency/style. 1=no shared details, 10=high detail overlap.
Be strict about concrete detail match.

Caption A:
{A}

Caption B:
{B}
\end{promptbox}
\newpage

\subsection{User Study to Verify LLM as a Judge for Metrics}
\begin{figure}
    \centering
    \includegraphics[width=1.0\linewidth]{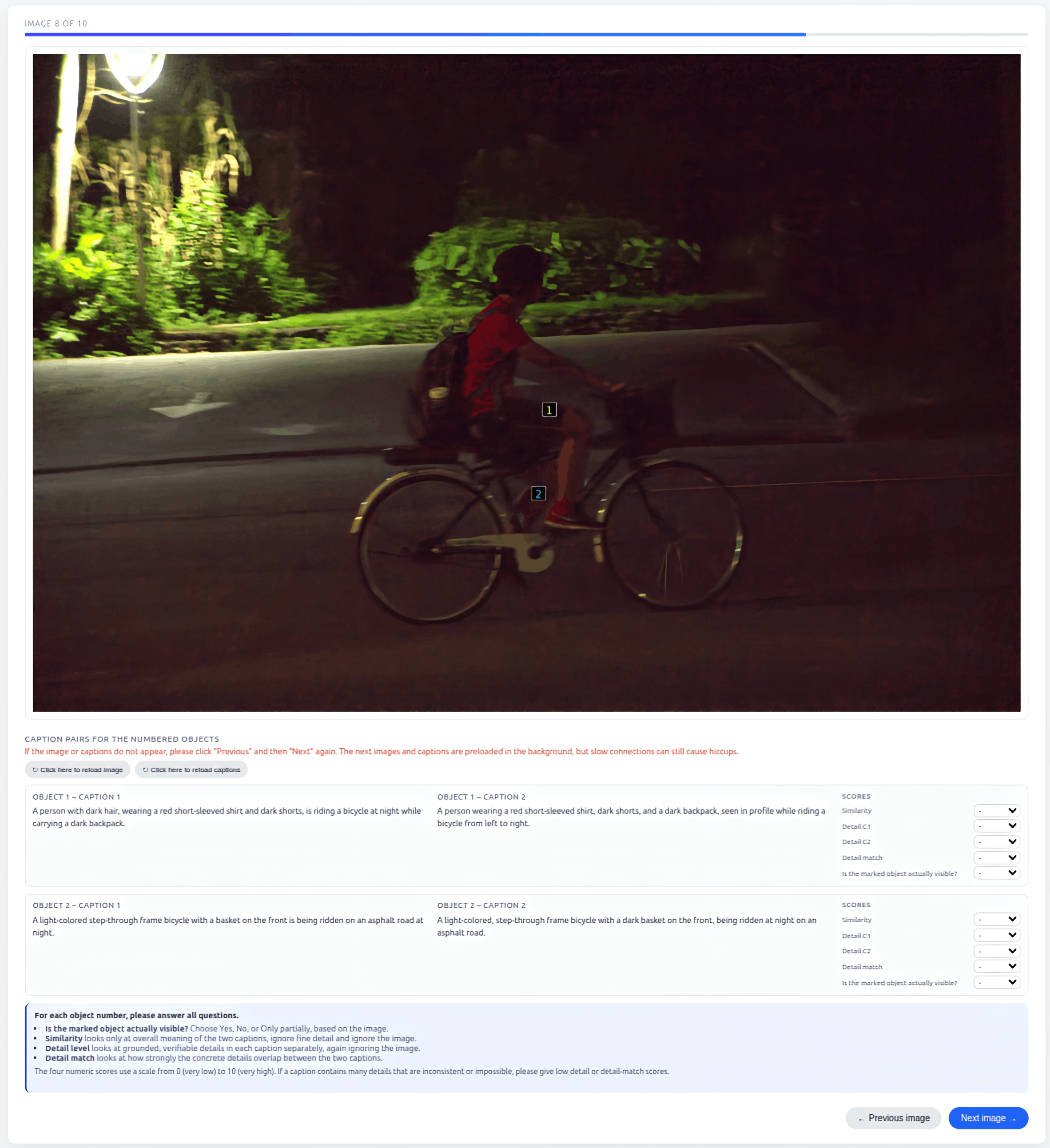}
    \caption{User-study example used to evaluate the performance of an LLM acting as a judge. Participants rated captions based on similarity with one another and richness of details. To assess the captioning model's hallucinations, they also indicated whether the objects described were visible in the images or not.}
    \label{fig:user_study_LLM}
\end{figure}

In order to validate the usage of LLMs as a judge to measure description similarity and level of detail, we also conducted a user study. \autoref{fig:user_study_LLM} provides an example from this user study. 
We gather a random subset of 50 images with the two object descriptions per marked object in the images, gathered from the high-resolution sRGB version of each image, and ask users to provide scores between 0 and 10  for the description similarity, the amount of details provided, and the agreement of the given details. 

Each user rates 10 of the 50 randomly sampled images at a time.
We collected 9 such user studies, so 90 images and their respective pair of object descriptions for each marked object were rated, that is, a total of 391 caption pairs. 
We calculate the correlation in the mean similarity, mean detail level, and mean detail match between Gemini-2.5-Flash-Lite as a judge and a human as a judge from each user study.
For similarity, we get a correlation of 0.61, for the detail level, we get a correlation of 0.71, and for the detail match, we get a correlation of 0.58. 
There are all high positive correlations, demonstrating that Gemini-2.5-Flash-Lite can be used as a judge for comparing two object descriptions at a time. 
Please note, using a better LLM like ChatGPT5 or Gemini-2.5-Pro as a judge might lead to better correlations with humans as a judge; however, this would incur high costs, and thus using Gemini-2.5-Flash-Lite is the most cost-effective way. 
Using open-source models for LLM as a judge might seem like a reasonable choice; however, as shown by \cite{agnihotri2025a}, open-source models like Qwen3, GPT-oss have significantly low correlation with human judgment compared to closed and paid models.

The specific instructions given to the user, for ``human as a judge'' are:\\

\noindent\footnotesize\fbox{%
    \parbox{0.95\textwidth}{
 Thank you for taking part in this study. In each trial you will see a high-resolution image with numbered marks on several objects. For every numbered object there are two automatically generated captions that are intended to describe the same marked object.

Your task is to rate the relationship between the two captions for each numbered object. For every object number, you will provide one visibility judgment and four scores on a scale from 0 to 10:
\begin{itemize}
\item Is the marked object actually visible? (Dependent on the image) Decide whether the object with this number is fully visible in the image, only partially visible, or not visible at all.

\item Semantic similarity (Independent of the image) - Do both captions clearly talk about the same object, ignoring fine-grained detail.\\

\item Level of detail of Caption 1 (Independent of the image) - How rich and specific Caption 1 is, in terms of concrete, verifiable details (independent of the image, verifying if the caption itself has obvious conflicts in the text).\\
    
\item Level of detail of Caption 2 (Independent of the image) - Same as above, now for Caption 2.\\

\item Detail match (Independent of the image) - How well the concrete details in Caption 1 and Caption 2 agree with each other.\\
\end{itemize}

Rating scale (0 to 10)
\begin{itemize}
    \item 0 - No similarity or no grounded details at all.
    \item 5 - Moderate similarity or a moderate amount of grounded details.
    \item10 - Very high similarity or very rich and specific grounded details.
\end{itemize}

The 0 to 10 scale applies to the four numeric scores. The visibility question instead uses the options Yes, No, and Only partially.\\

Only for the visibility question (Is the marked object actually visible?), you should look at the image. For all other scores, please ignore the image itself and judge only the two captions for that object. If a caption contains many details that are internally inconsistent or impossible (for example, mentioning both headlights and taillights when only the front of a car can be visible), please give low scores for the relevant detail and detail-match questions.\\

Please pay attention that each caption is tied to a numbered object in the image. When you score, always think about the object with that number, not the whole image. 
}}

\end{document}